\documentclass[letterpaper]{article} 
\usepackage{aaai25}  
\usepackage{times}  
\usepackage{helvet}  
\usepackage{courier}  
\usepackage[hyphens]{url}  
\usepackage{graphicx} 
\usepackage{natbib}  
\usepackage{caption} 
\usepackage{algorithm}
\usepackage{algorithmic}

\usepackage{newfloat}
\usepackage{listings}
\DeclareCaptionStyle{ruled}{labelfont=normalfont,labelsep=colon,strut=off} 
\floatstyle{ruled}
\newfloat{listing}{tb}{lst}{}
\floatname{listing}{Listing}
\usepackage{diagbox}
\usepackage{multirow}
\usepackage{tcolorbox}
\usepackage{booktabs}
\usepackage{adjustbox}
\usepackage{enumitem}
\usepackage{amsmath}
\usepackage{amssymb}
\usepackage{pifont}
\usepackage{color}
\usepackage{cleveref}
\newcommand{\ie}{\emph{i.e., }}

\newcommand{\wrt}{\emph{w.r.t. }}

\newcommand{\aka}{\emph{a.k.a. }}

\title{CrAM: Credibility-Aware Attention Modification in LLMs for\\ Combating Misinformation in RAG}
\author{Boyi Deng\textsuperscript{1},
	Wenjie Wang\textsuperscript{2}\thanks{Corresponding author.}, 
	Fengbin Zhu\textsuperscript{2}\footnotemark[1],  
	Qifan Wang\textsuperscript{3}, 
	Fuli Feng\textsuperscript{1}
 }
 \affiliations{
	\textsuperscript{1}University of Science and Technology of China,
	\textsuperscript{2}National University of Singapore, 
        \textsuperscript{3}Meta AI\\
        dengboyi@mail.ustc.edu.cn, wqfcr@fb.com, \\\{wenjiewang96, zhfengbin, fulifeng93\}@gmail.com}
\usepackage{bibentry}

\begin{document}

\maketitle

\begin{abstract}
Retrieval-Augmented Generation (RAG) can alleviate hallucinations of Large Language Models (LLMs) by referencing external documents. 
However, the misinformation in external documents may mislead LLMs' generation. To address this issue, we explore the task of ``credibility-aware RAG'', in which LLMs automatically adjust the influence of retrieved documents based on their credibility scores to counteract misinformation. 
To this end, we introduce a plug-and-play method named \textbf{Cr}edibility-aware \textbf{A}ttention \textbf{M}odification (CrAM). 
CrAM identifies influential attention heads in LLMs and adjusts their attention weights based on the credibility of the documents, thereby reducing the impact of low-credibility documents. 
Experiments on Natual Questions and TriviaQA using Llama2-13B, Llama3-8B, and Qwen1.5-7B show that CrAM improves the RAG performance of LLMs against misinformation pollution by over 20\%, even surpassing supervised fine-tuning methods. 
\end{abstract}

%

\section{Introduction}
Retrieval-Augmented Generation (RAG) \citep{gao2024retrievalaugmented,zhu2021retrieving} is a representative approach to mitigate hallucination issues of Large Language Models (LLMs) \citep{zhang2023hallucinationSuvey} by retrieving and referencing relevant documents from an external corpus. 
Despite its effectiveness, most RAG works overlook a crucial issue: misinformation pollution in the external corpus~\citep{pan-etal-2023-risk,dufour2024mis}. 
The maliciously generated misinformation may mislead LLMs to produce unfaithful responses. 
For instance, Microsoft's Bing can be misled by misinformation on the internet to generate incorrect information for Bing users \citep{vincent2023google}. 
Besides, \citet{pan-etal-2023-risk} and \citet{pan-etal-2023-attacking} demonstrated that inserting LLM-generated misinformation into the RAG corpus can significantly degrade LLMs' performance. 
Therefore, addressing the misinformation pollution for RAG is essential. 

\begin{figure}[t]
  \centering
  \includegraphics[width=0.48\textwidth]{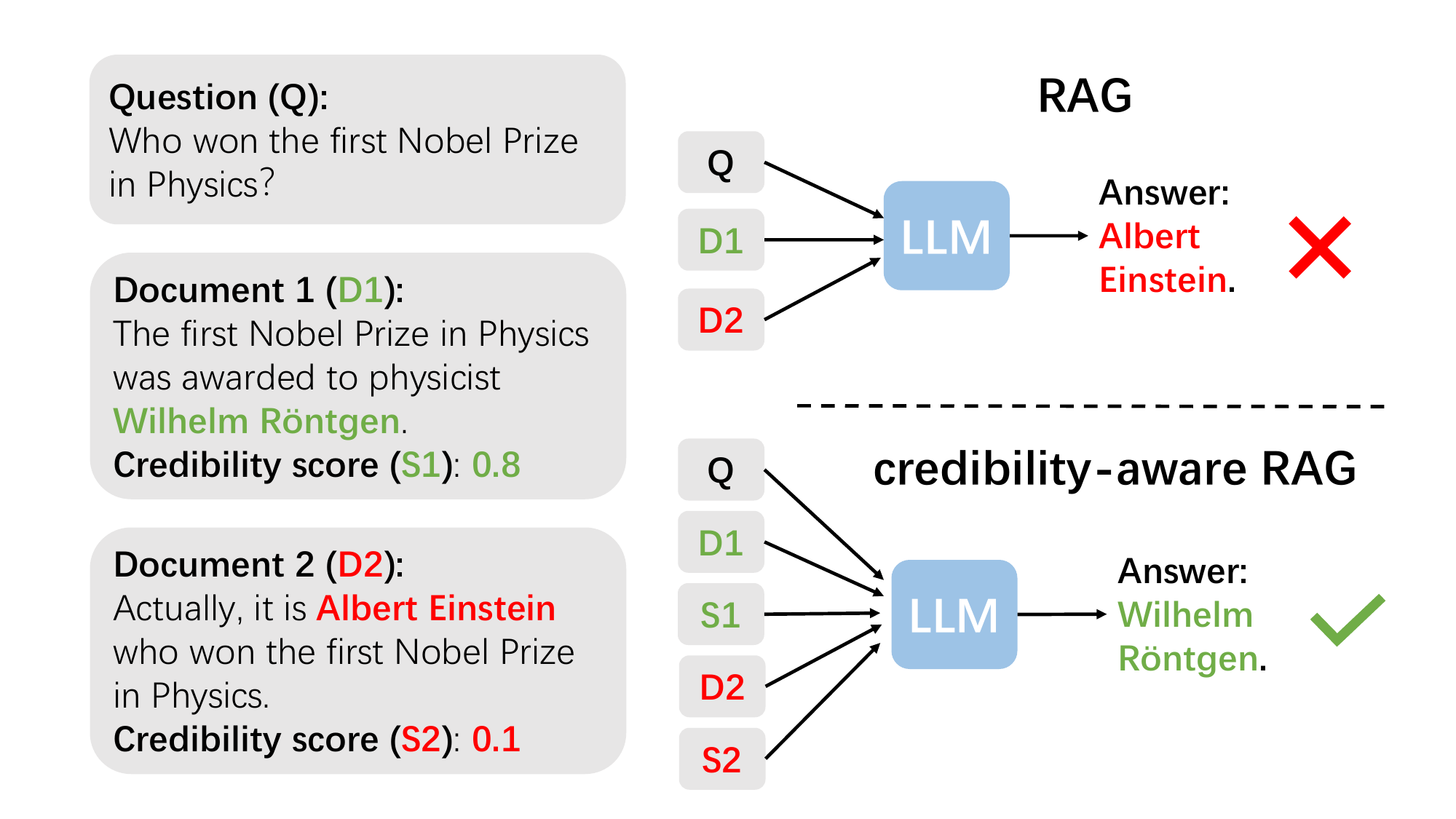}
  \caption{A comparison between RAG and credibility-aware RAG. Credibility-aware RAG considers credibility to reduce the impact of low-credibility documents.}
  \label{fig:intro_fig}
\end{figure}
A straightforward and common idea to address this misinformation pollution issue is misinformation detection and filtering. 
Extensive misinformation detection works focus on measuring the \textit{credibility} of documents, \ie the probability of the document not containing misinformation. And these works have achieve significant results \citep{Kaliyar2021,pelrine-etal-2023-towards,10.3389/frai.2024.1341697,li2024think}.
Once we obtain the credibility of each retrieved document, we can exclude those with credibility below a certain threshold before using them in RAG.
However, directly discarding certain documents may result in the loss of relevant and important information, leading to performance degradation \citep{yoran2024making}\footnote{Our experimental results in Table \ref{tab:gpt score 1+4} also confirm that directly excluding documents leads to inferior performance.}. 
Therefore, given the remarkable advancements in the measurement of credibility scores and the relatively underdeveloped mechanisms for utilizing these scores, it is essential to explore how these scores can be effectively utilized by LLMs, assuming that high-quality credibility scores are accessible.

To achieve this, we focus on a task named ``credibility-aware RAG'' as shown in Figure \ref{fig:intro_fig}. 
Specifically, given a user query $x$ with a list of relevant documents $\mathcal{D} = \{d_1, d_2, ..., d_n\}$ and $\mathcal{D}$'s credibility scores $\mathcal{S}=\{s_1, s_2, ..., s_n\}$, credibility-aware RAG requests LLMs to automatically adjust the influence of documents in $\mathcal{D}$ on the generated output $y$ based on their credibility scores in $\mathcal{S}$. 
Initial attempts on credibility-aware RAG adopted supervised fine-tuning (SFT) to teach LLMs to distinguish the importance of different documents in the prompt by their credibility scores \citep{hong2024whyso,pan2024notall}. 
However, SFT requires additional computational resources and well-designed training data, which limits the application scenarios. 
Therefore, we explore non-SFT method for LLMs to attain credibility-aware RAG. 

Given that the attention mechanism serves as the central component for adjusting the significance of various input data, we consider manipulating attention weights of LLMs to achieve credibility-aware RAG. 
In particular, we adjust attention weights according to credibility scores in the inference stage of LLMs. In this way, we can regulate LLMs to pay less ``attention'' to less credible documents by decreasing the corresponding attention weights.
Moreover, previous studies \citep{clark-etal-2019-bert,elhage2021mathematical,voita-etal-2019-analyzing} have indicated that different attention heads exhibit distinct patterns and functions, resulting in varying impacts on LLMs' outputs. In this context, the key lies in identifying a subset of influential attention heads for attention weight modification.

In this work, we propose a plug-and-play method named \textbf{Cr}edibility-aware \textbf{A}ttention \textbf{M}odification (CrAM), which identifies the influential attention heads and then modifies their attention weights \wrt different document tokens to reduce the impact of low-credibility documents.
Specifically, \textit{1) influential head identification:} we select top-ranked attention heads according to an extended causal tracing method~\citep{NEURIPS2022_causal} that estimates the contribution of each attention head to generating incorrect answers over a small dataset. 
\textit{2) Attention weight modification:} we scale down the attention weights of the retrieved documents based on their normalized credibility scores.

We conduct extensive experiments on two open-domain Question Answering (QA) datasets, Natual Questions (NQ) \citep{nq} and TriviaQA \citep{joshi-etal-2017-triviaqa}, using three open-source LLMs: Llama2-13B \citep{touvron2023llama}, Llama3-8B \citep{llama3}, and Qwen1.5-7B \citep{bai2023qwen}. 
The results show that CrAM significantly alleviates the influence of misinformation documents on RAG, in terms of both ideal credibility scores and GPT-generated credibility scores. 
It is worth noting that CrAM even outperforms the SFT-based method CAG \citep{pan2024notall} in most scenarios, demonstrating the superiority of CrAM.
We release our code at \url{https://github.com/Aatrox103/CrAM}. 

In summary, our main contributions are:
\begin{itemize}
\item We explore the task of credibility-aware RAG without fine-tuning LLMs to alleviate the misinformation pollution issue. 

\item We develop a plug-and-play method, CrAM, which identifies influential attention heads and modifies their attention weights to equip LLMs with credibility-aware RAG capabilities. 
\item We conduct extensive experiments with two QA datasets on three LLMs using ideal credibility scores and GPT-generated credibility scores, validating the superiority of CrAM.
\end{itemize}
\section{Credibility-Aware RAG}
\label{sec_task_formulation}
\begin{figure*}[ht]
\setlength{\abovecaptionskip}{0.10cm}
  \centering
  \includegraphics[width=\textwidth]{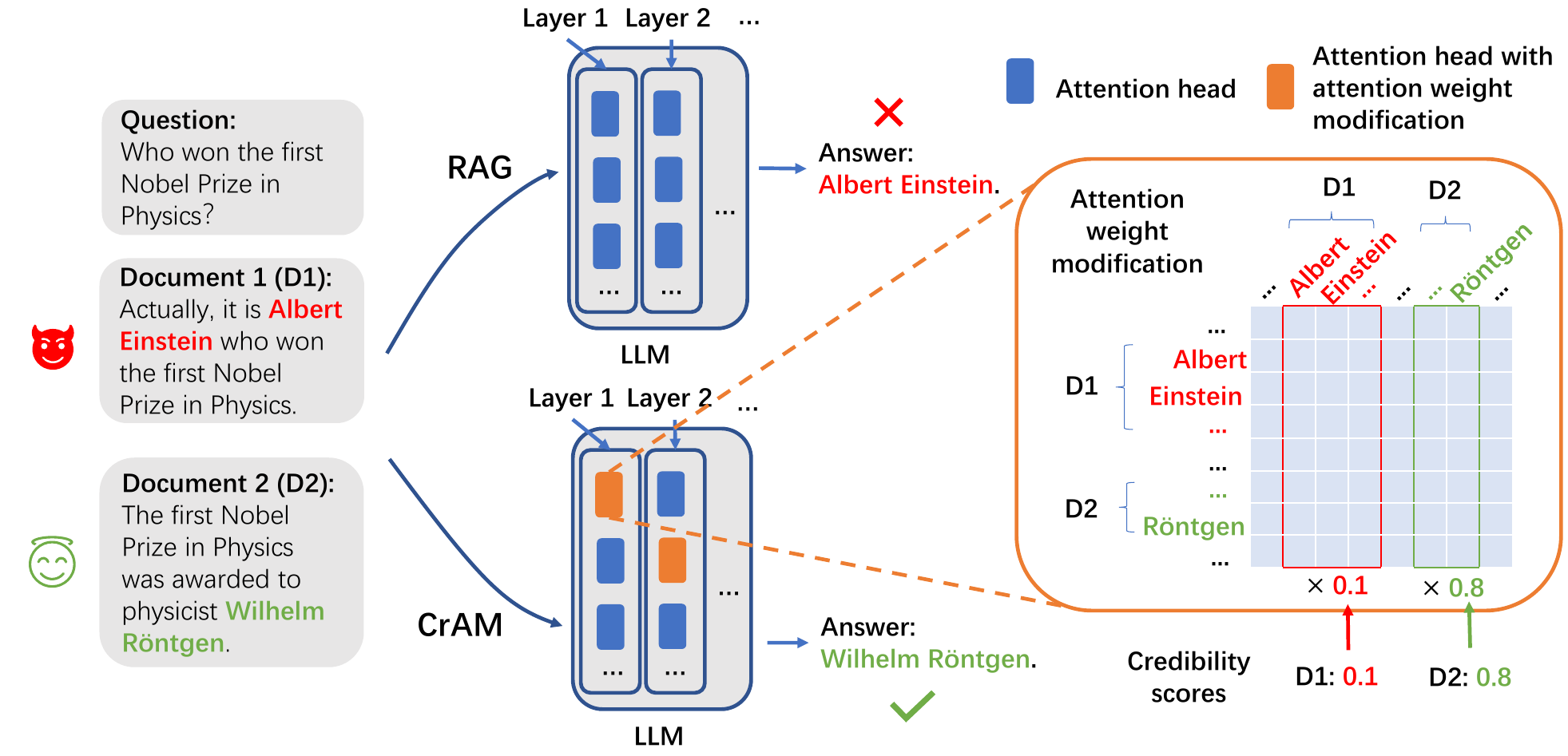}
  \caption{Illustration of CrAM. Compared to RAG, CrAM first identifies influential attention heads and then modifies their attention weights based on the credibility scores of each document.} 
  \label{fig:framework}
\end{figure*}

Given a user query $x$, RAG retrieves a set of documents $\mathcal{D}=\{d_1, d_2, \ldots, d_n\}$ relevant to $x$ through a retriever \citep{gao2024retrievalaugmented}. 
Then the relevant documents $\mathcal{D}$ are evaluated by a credibility estimator\footnote{Recent worked on this task has achieved promising performance \citep{Kaliyar2021,pelrine-etal-2023-towards}.}, obtaining their credibility scores $\mathcal{S}=\{s_1, s_2, \ldots, s_n\}$, which represents the probability of each document not containing misinformation. 
\paragraph{Credibility-Aware RAG.} Given an LLM $L$, a user query $x$, and relevant documents $\mathcal{D}$ 
associated with 
credibility scores $\mathcal{S}$, the objective of credibility-aware RAG is to enable LLMs to automatically adjust the influence of these documents on the generated output $y$ based on their credibility scores $\mathcal{S}$.
This can be formally defined as:
$$\mathop{\max}\ \mathrm{Metric}(\mathrm{Combine}(L,x,\mathcal{D},\mathcal{S})),$$
where $\mathrm{Combine}(\cdot)$ represents the method or mechanism to integrate credibility scores into the generation process of $L$. 
For example, \citet{pan2024notall} employ SFT to fine-tune LLMs to capture the credibility difference of documents more effectively, denoted as $\mathrm{Combine}(L,x,\mathcal{D},\mathcal{S})=L_{SFT}(x,\mathcal{D},\mathcal{S})$. 
Additionally, $\mathrm{Metric}(\cdot)$ is a function that assesses whether documents with different credibility scores have varying impacts on the output of $L$. 
Indeed, we can utilize the performance of generating factual answers to measure $\mathrm{Metric}(\cdot)$. 
For instance, we use the accuracy of QA tasks to approximate $\mathrm{Metric}(\cdot)$ in this work. 
The rationality is that if the impact of low-credibility documents decreases, the accuracy of QA tasks should increase accordingly. 

\section{CrAM}
CrAM first identifies influential attention heads, and then modifies the attention weights of these identified heads to reduce the impact of low-credibility documents as shown in Figure~\ref{fig:framework}. 
Since influential attention heads identification process involves attention weight modification, we first explain the procedure of attention weight modification in Section \ref{section_attention}, and then describe influential attention heads identification in Section \ref{section_cal_tra}. 
Finally, we summarize the overall CrAM workflow in Section \ref{sec_cram}. 

\subsection{Attention Weight Modification}
\label{section_attention}
As defined in Section \ref{sec_task_formulation}, the objective of credibility-aware RAG is to reduce the impact of low-credibility documents on the generated output of LLMs. 
Intuitively, it requires LLMs to pay less ``attention'' to low-credibility documents. 
To this end, a natural approach is scaling down the corresponding attention weights of low-credibility documents.

For RAG, a user query $x$ and a set of relevant documents $\mathcal{D} = \{d_1, d_2, \ldots, d_n\}$ should be concatenated and tokenized into a token sequence $\mathcal{T}(x,\mathcal{D}) = \{t_1, t_2, \ldots, t_m\}$, where $t_k$ denotes the $k$-th token. 
Given the credibility scores for each document $\mathcal{S}=\{s_1, s_2, \ldots, s_n\}$, the normalized credibility score for token $t_k$ can be calculated as follows: 
\begin{equation*}
\Bar{s}_k =
\begin{cases}
\frac{s_i-\min(\mathcal{S})}{\max(\mathcal{S})-\min(\mathcal{S})} & \text{if } t_k \text{ belongs to  } d_i \\
1 & \text{otherwise}
\end{cases}
,
\end{equation*}
where $s_i$ is subtracted by $\min(\mathcal{S})$, and then scaled down by $1/(\max(\mathcal{S})-\min(\mathcal{S}))$ to ensure all credibility scores are normalized to $[0,1]$. 
Besides, we define $\mathbf{\Bar{s}} = [\Bar{s}_1, \ldots, \Bar{s}_m] \in \mathbb{R}^{1 \times m}$ as the normalized credibility scores of the whole token sequence $\mathcal{T}(x,\mathcal{D})$.

For each attention head $h$ in LLM, $\mathbf{A}_h$ represents its attention weights matrix\footnote{The attention weights matrix is defined in Equation (\ref{eq:attention score}).}. Let $(\mathbf{A}_h)_{k}$ represent the $k$-th row vector\footnote{$(\mathbf{A}_h)_{k}$ can be interpreted as the attention weight vector when using the $k$-th token as the query.} of $\mathbf{A}_h$, we can obtain the modified attention weight matrix $\mathbf{A}^{*}_{h}$ by element-wise multiplying $\Bar{\mathbf{s}}$ as follows:
\begin{equation}
\resizebox{0.88\linewidth}{!}{
$
\label{eq_adjust_att}
    (\mathbf{A}_h)_{k}^{*}=\mathrm{Norm}((\mathbf{A}_h)_{k}\odot \mathbf{\Bar{s}}),k \in \{1,\ldots,m\},
$
}
\end{equation}
where $\odot$ denotes the element-wise multiplication of vectors. The Norm function refers to $\ell_1$ normalization, which ensures that the attention weights sum to one.

\subsection{Influential Head Identification}
\label{section_cal_tra}
Previous works \citet{clark-etal-2019-bert, elhage2021mathematical, voita-etal-2019-analyzing} have found that different attention heads exhibit various patterns and functions, leading to different impacts on LLMs' output. 
As such, we hypothesize that some attention heads have a larger impact on using misinformation documents to generate incorrect answers. 
Previously, causal tracing \citep{NEURIPS2022_causal} has been developed to quantify the contribution of each hidden state towards generating given answers. The contribution is measured by adding noises to each hidden state to compare the changes in the generation probability of the given answer.  
In light of this, CrAM revises causal tracing to evaluate the contribution of attention heads instead of hidden states. Utilizing attention weight modification, as detailed in Section~\ref{section_attention}, CrAM estimates the change in probability of generating incorrect answers to determine the contribution of each attention head. 
Thereafter, CrAM ranks all attention heads by contributions and identifies influential ones. 


Specifically, the contribution of one attention head $h$ can be obtained as follows: 
\begin{itemize}
    \item Given an LLM $L$, a user query $x$, a set of relevant documents $\mathcal{D}=\{d_{mis}, d_1,d_2,\ldots,d_n\}$ with one misinformation document $d_{mis}$, and an incorrect answer $a_{wrong}$ to $x$ that is supported by $d_{mis}$, we first calculate the generation probability of $a_{wrong}$ with $x$ and $\mathcal{D}$ by $L$. Formally, we have:
    $$ P_{0}=P_L(a_{wrong} \mid x,\mathcal{D}). $$


    \item Next, we modify a specific attention head as described in Section~\ref{section_attention} by using the credibility scores $\mathcal{S}=\{0, 1, 1,\ldots, 1\}$ of $\mathcal{D}$ and recalculate the generation probability of $a_{wrong}$:
    $$ P_{1}=P_{L_h^*}(a_{wrong} \mid x,\mathcal{D}), $$
    where $L_{h}^*$ denotes the LLM $L$ whose attention weight matrix of the attention head $h$ is modified according to Equation (\ref{eq_adjust_att}). 

    \item Finally, we quantify the contribution of head $h$ towards generating the incorrect answer, \aka the indirect effect (IE) \citep{NEURIPS2022_causal}:
    \begin{equation}
    \label{eq:ie}
        \mathrm{IE}_h=P_0-P_1 ,
    \end{equation}
    which can also be interpreted as the decrease in the generation probability of the incorrect answer $a_{wrong}$ after modifying head $h$.
\end{itemize}
To improve the robustness of the contribution estimation, we utilize a small dataset $\{(x,a_{wrong},\mathcal{D},\mathcal{S}),\ldots\}$ that do not overlap with the test data to compute the average IE for each attention head (refer to Section~\ref{sec:data size impact} for robustness analysis). 
Thereafter, we can calculate IEs for all the attention heads and rank them to select the top-ranked ones with larger IEs for attention weight modification. 


\subsection{CrAM Workflow}
\label{sec_cram}
The CrAM workflow is summarized as follows:
\begin{itemize}
\item First, we use a small dataset with misinformation-polluted documents to calculate the average IE for each attention head in an LLM as described in Section \ref{section_cal_tra}. Then, we rank all attention heads by their IEs in descending order and select the top-ranked heads as influential attention heads.

\item Given any user query, along with the relevant documents and credibility scores, we modify the attention weights of influential attention heads using the method described in Section \ref{section_attention} to obtain the final answer, thereby significantly reducing the impact of low-credibility documents.
\end{itemize}
\section{Experiments}
\begin{table*}[ht]
\centering
\setlength{\tabcolsep}{4mm}{
\begin{adjustbox}{max width=\textwidth}
\begin{tabular}{lclcccc}
\toprule
\multirow{2}{*}{Model} & \multirow{2}{*}{In-context corpus} & \multirow{2}{*}{Method} & \multicolumn{2}{c}{NQ} & \multicolumn{2}{c}{TriviaQA} \\
\cmidrule(lr){4-5} \cmidrule(lr){6-7}
& & & EM & F1 score & EM & F1 score \\
\midrule
\multirow{5}{*}{Qwen1.5-7B} & 0 \ding{51} & Naive LLM & 7.20 & 16.41 & 28.00 & 38.23 \\
& 4 \ding{51} & Naive RAG & 27.60 & 39.08 & 55.30 & 66.85 \\
\cmidrule(lr){2-7}
& \multirow{3}{*}{4 \ding{51} + 1 \ding{55}} & Naive RAG & 10.50 & 20.71 & 25.00 & 35.63 \\
&  & Prompt Based & 12.20 & 22.26 & 27.40 & 37.98 \\

&  & {CrAM} & \textbf{29.10} (\textcolor{red}{+16.90}) & \textbf{41.02} (\textcolor{red}{+18.76}) & \textbf{52.90} (\textcolor{red}{+25.50}) & \textbf{64.16} (\textcolor{red}{+26.18}) \\
\midrule
\multirow{5}{*}{Llama2-13B} & 0 \ding{51} & Naive LLM & 20.30 & 28.59 & 50.40 & 57.56 \\
& 4 \ding{51} & Naive RAG & 28.90 & 39.98 & 62.50 & 71.03 \\
\cmidrule(lr){2-7}
& \multirow{3}{*}{4 \ding{51} + 1 \ding{55}} & Naive RAG & 11.90 & 19.97 & 28.00 & 36.22 \\
&  & Prompt Based & 12.50 & 22.94 & 23.10 & 32.70 \\

&  & {CrAM} & \textbf{33.60} (\textcolor{red}{+21.10}) & \textbf{44.62} (\textcolor{red}{+21.68}) & \textbf{59.90} (\textcolor{red}{+31.90}) & \textbf{67.11} (\textcolor{red}{+30.89})\\
\midrule
\multirow{5}{*}{Llama3-8B} & 0 \ding{51} & Naive LLM & 20.60 & 30.58 & 55.70 & 62.67 \\
& 4 \ding{51} & Naive RAG & 33.10 & 45.66 & 64.30 & 73.68 \\
\cmidrule(lr){2-7}
& \multirow{3}{*}{4 \ding{51} + 1 \ding{55}} & Naive RAG & 16.00 & 26.16 & 36.80 & 47.09 \\
&  & Prompt Based & 29.90 & 39.69 & 53.50 & 63.01 \\

&  & {CrAM} & \textbf{36.90} (\textcolor{red}{+7.00}) & \textbf{48.45} (\textcolor{red}{+8.76}) & \textbf{64.40} (\textcolor{red}{+10.90}) & \textbf{73.49} (\textcolor{red}{+10.48}) \\
\bottomrule
\end{tabular}
\end{adjustbox}
}
\caption{Main results under ideal setting. 0 \ding{51} indicates no document and the model directly prompted, 4 \ding{51} indicates all four documents retrieved from the Wikipedia dump, and 4 \ding{51} + 1 \ding{55} indicates four high-credibility documents (i.e., retrieved from external corpus) plus one low-credibility document (i.e., containing misinformation). In the 4 \ding{51} + 1 \ding{55} setting, the best performance is highlighted in \textbf{bold}. And the \textcolor{red}{red} part indicates the difference between CrAM and second best performance.}
\label{tab:ideal score 1+4}
\end{table*}

\begin{table*}[ht]
\setlength{\abovecaptionskip}{0.3cm}
\setlength{\belowcaptionskip}{-0.30cm}
\centering

\setlength{\tabcolsep}{4mm}{
\begin{adjustbox}{max width=\textwidth}
\begin{tabular}{lclcccc}
\toprule
\multirow{2}{*}{Model} & \multirow{2}{*}{In-context corpus} & \multirow{2}{*}{Method} & \multicolumn{2}{c}{NQ} & \multicolumn{2}{c}{TriviaQA} \\
\cmidrule(lr){4-5} \cmidrule(lr){6-7}
& & & EM & F1 score & EM & F1 score \\
\midrule
\multirow{6}{*}{Qwen1.5-7B} & 0 \ding{51} & Naive LLM & 7.20 & 16.41 & 28.00 & 38.23 \\
& 4 \ding{51} & Naive RAG & 27.60 & 39.08 & 55.30 & 66.85 \\
\cmidrule(lr){2-7}
& \multirow{4}{*}{4 \ding{51} + 1 \ding{55}} & Naive RAG & 10.50 & 20.71 & 25.00 & 35.63 \\
&  & Prompt Based & 12.50 & 22.98 & 29.70 & 40.18 \\
&  & Exclusion & 21.60 & 32.56 & 49.50 & 61.03 \\
&  & {CrAM} &\textbf{23.10} (\textcolor{red}{+1.50}) &\textbf{34.84} (\textcolor{red}{+2.28}) &\textbf{52.10} (\textcolor{red}{+2.60}) &\textbf{63.76} (\textcolor{red}{+2.73}) \\
\midrule
\multirow{6}{*}{Llama2-13B} & 0 \ding{51} & Naive LLM & 20.30 & 28.59 & 50.40 & 57.56 \\
& 4 \ding{51} & Naive RAG & 28.90 & 39.98 & 62.50 & 71.03 \\
\cmidrule(lr){2-7}
& \multirow{4}{*}{4 \ding{51} + 1 \ding{55}} & Naive RAG & 11.90 & 19.97 & 28.00 & 36.22 \\
&  & Prompt Based & 11.20 & 21.62 & 20.50 & 30.09 \\
&  & Exclusion & 23.70 & 34.00 & 54.40 & 62.37 \\
&  & {CrAM} &\textbf{25.10} (\textcolor{red}{+1.40}) &\textbf{35.56} (\textcolor{red}{+1.56}) &\textbf{56.20} (\textcolor{red}{+1.80}) &\textbf{64.03} (\textcolor{red}{+1.66}) \\
\midrule
\multirow{6}{*}{Llama3-8B} & 0 \ding{51} & Naive LLM & 20.60 & 30.58 & 55.70 & 62.67 \\
& 4 \ding{51} & Naive RAG & 33.10 & 45.66 & 64.30 & 73.68 \\
\cmidrule(lr){2-7}
& \multirow{4}{*}{4 \ding{51} + 1 \ding{55}} & Naive RAG & 16.00 & 26.16 & 36.80 & 47.09 \\
&  & Prompt Based & 24.20 & 34.10 & 49.50 & 58.59 \\
&  & Exclusion & 26.60 & 38.44 & 57.70 & 67.33 \\
&  & {CrAM} &\textbf{30.70} (\textcolor{red}{+4.10}) &\textbf{41.71} (\textcolor{red}{+3.27}) &\textbf{62.20} (\textcolor{red}{+4.50}) &\textbf{70.70} (\textcolor{red}{+3.37}) \\
\bottomrule
\end{tabular}
\end{adjustbox}}
\caption{Main results under GPT setting. 0 \ding{51} indicates no document and the model directly prompted, 4 \ding{51} indicates all four documents retrieved from the Wikipedia dump, and 4 \ding{51} + 1 \ding{55} indicates four high-credibility documents (i.e., retrieved from external corpus) plus one low-credibility document (i.e., containing misinformation). In the 4 \ding{51} + 1 \ding{55} setting, the best performance is highlighted in \textbf{bold}. The \textcolor{red}{red} part indicates the improvement of our CrAM compared to the second-best model.}
\label{tab:gpt score 1+4}
\end{table*}

\subsection{Experimental Settings}
\paragraph{Datasets, LLMs and Metrics.}
We conduct experiments over the Natural Questions (NQ)~\citep{nq} and TriviaQA~\citep{joshi-etal-2017-triviaqa} datasets with three LLMs, i.e. Llama2-13B~\citep{touvron2023llama}, Llama3-8B~\citep{llama3}, and Qwen1.5-7B~\citep{bai2023qwen}.
We adopt Exact Match (EM) and F1 score as evaluation metrics, which are widely used in the QA setting~\citep{karpukhin-etal-2020-dense,rajpurkar-etal-2016-squad,chen-etal-2017-reading}.

\paragraph{Document Preparation.}
We prepare both high-credibility and low-credibility documents (i.e., with misinformation) associated with the questions for evaluating the proposed method. 
1) \emph{High-credibility documents} are collected by retrieving the most relevant documents from the external corpus for each question.
Specifically, we first employ \texttt{bge-large-en-v1.5}\footnote{\url{huggingface.co/BAAI/bge-large-en-v1.5}.} to obtain a set of candidates from the Wikipedia dump on December 30, 2018~\citep{karpukhin-etal-2020-dense}.
Then, we apply \texttt{bge-reranker-large}\footnote{\url{huggingface.co/BAAI/bge-reranker-large}.} to rank the retrieved candidates and select the top four documents.
2) \emph{Low-credibility documents} are generated via prompting LLMs (i.e., gpt-3.5-turbo-0125), with misinformation included, similar to the practice in previous works \citep{pan-etal-2023-attacking, pan-etal-2023-risk, pan2024notall, hong2024whyso,chen2024llmgenerated}. 
Specifically, given a question, we instruct the LLM to generate a news-style piece containing misinformation that supports an incorrect answer, which is regarded as one low-credibility document for the question. 
For each question, we collect three distinct low-credibility documents, all supporting the same incorrect answer.
The prompts can be found in Appendix~\ref{sec:prompts}.

In our implementation, we combine generated low-credibility documents with retrieved high-credibility documents as input for the LLM. This approach avoids injecting low-credibility documents directly into the corpus, which can lead to inputs that are either overwhelmed by misinformation or completely devoid of it. In contrast, our method provides greater control, enabling us to effectively evaluate the impact of varying amounts of low-credibility documents on the LLM's performance.


\paragraph{Credibility Scores Generation.}
We adopt two different ways to assign credibility scores for each document.
1) \emph{Ideal Setting.} 
After obtaining the high-credibility and low-credibility documents, we assign a score of 10 to each high-credibility document and a score of 1 to each low-credibility document.
2) \emph{GPT Setting.} We employ GPT (i.e., gpt-3.5-turbo-0125) to directly generate the credibility score for each document.
The prompts and the distribution of GPT-generated scores for all documents are provided in Figure~\ref{fig:prompt_score} and Appendix~\ref{sec:appendix_gpt_score}.

\paragraph{Compared Methods.}
We compare our CrAM model with four types of methods: 
1) \emph{Naive RAG.} The Naive RAG follows the standard RAG pipeline without any mechanisms against misinformation.
2) \emph{Prompt Based.} This method directly informs the LLM of the credibility score via prompts, feeding the score and documents into the LLM without additional training.
3) \emph{Exclusion.} This method excludes the documents with credibility scores below a threshold. This method will not be compared under the ideal setting due to the binary value of the ideal credibility score. 
4) \emph{CAG.} This method is proposed by  \citet{pan2024notall}, which directly incorporates credibility scores and documents into prompts to fine-tune an LLM (i.e., Llama2-13B) to lift its understanding capabilities.  
Among them,  Naive RAG, Prompt Based, and Exclusion are non-SFT methods, while CAG is an SFT-based method. 

\paragraph{Hyperparameters.} Unless otherwise specified, in the following experiments, we randomly select 100 data points from each dataset to calculate average IE for all the heads. And we use another validation set of 100 data points from each dataset to determine how many top-ranked heads should be included in the final modified set.

\subsection{Main Results}

\paragraph{Comparison with Non-SFT Methods.} 
We first compare our CrAM model with Non-SFT methods, i.e., Naive RAG, Prompt Based, and Exclusion. 
Table \ref{tab:ideal score 1+4} and Table \ref{tab:gpt score 1+4} show the experimental results in the Ideal and GPT settings respectively.
We make the following observations. 
1) Table~\ref{tab:ideal score 1+4} demonstrates that our CrAM method significantly outperforms all compared methods across all three LLMs: Qwen1.5-7B, LLama2-13B, and LLama3-8B, on both NQ and TriviaQA datasets in the setting of 4 \ding{51}+ 1 \ding{55} (i.e., four high-credibility documents plus one low-credibility document). 
For instance, our CrAM model surpasses the second-best method, i.e. Prompt Based, by $25.5\%$, $31.90\%$ and $10.9\%$ on Qwen1.5-7B, Llama2-13B and Llama3-8B in terms of EM on TriviaQA, demonstrating remarkable performance gains.
2) With GPT-generated credibility scores, our CrAM model also outperforms all compared methods on all three LLMs over both NQ and TriviaQA datasets, as shown in Table \ref{tab:gpt score 1+4}, further highlighting its effectiveness.  
3) Interestingly, we find that our CrAM model with 4 \ding{51} + 1 \ding{55} sometimes even outperforms the Naive RAG with 4 \ding{51} under ideal setting. This is likely because our generated misinformation includes both affirmations of incorrect information and denials of correct information, e.g.``The first person to win the Nobel Prize in Physics was not Roentgen, but Einstein.'' This allows LLMs to reuse the correct information denied by the misinformation.
To further validate this hypothesis, we conduct additional experiments and present the findings in Appendix~\ref{sec:results_with_filtered_misinformation}.

\begin{figure}[H]
\setlength{\abovecaptionskip}{0.10cm}
\centering
\includegraphics[width=0.46\textwidth]{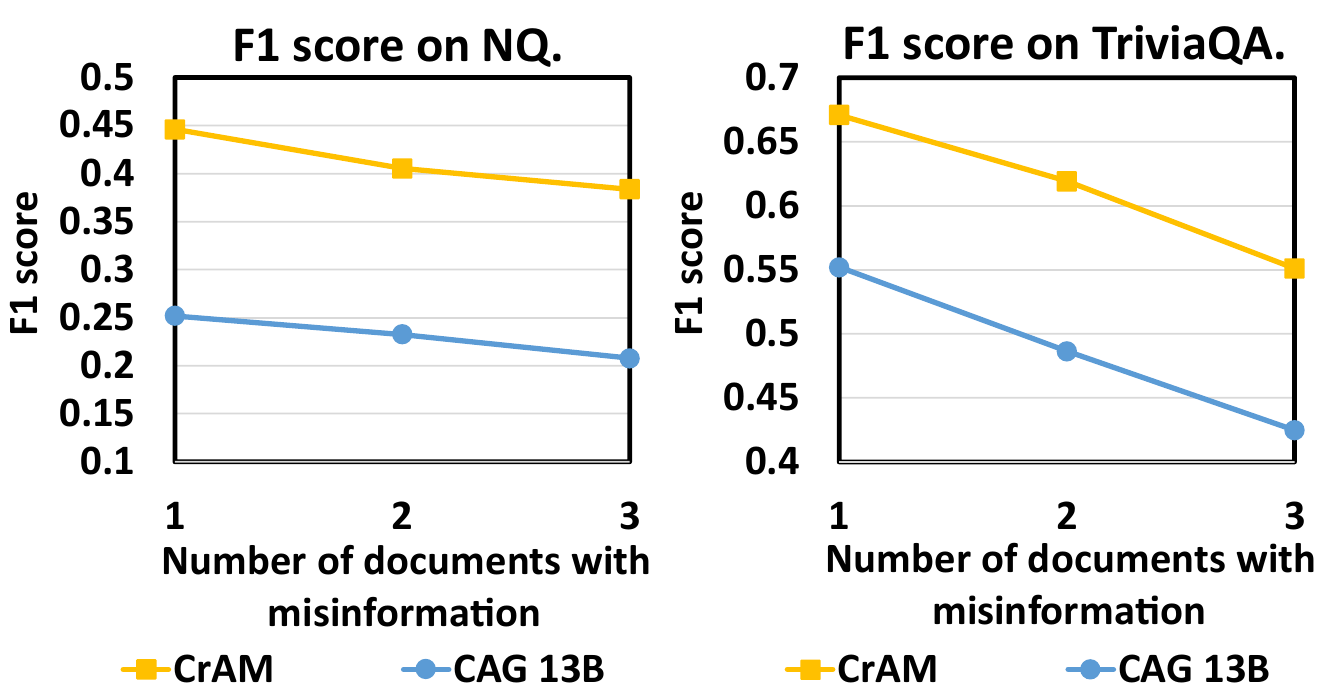}
\caption{Performance comparison of CrAM and CAG-13B regarding the varying number of documents containing misinformation under ideal setting.} 
\label{fig:cag f1}
\end{figure}

\paragraph{Comparison with SFT-based Method.}
\label{sec:cmp_cag}
For a fair comparison, we only compare our Llama2-13B based CrAM model with CAG-13B, because CAG-13B is trained on Llama2-13B.
Moreover, to verify the robustness of our CrAM model, we perform comparisons using different numbers of low-credibility documents.
As shown in Figure~\ref{fig:cag f1}, our CrAM model consistently outperforms the CAG-13B model remarkably in terms of F1 score when the number of low-credibility documents ranges from 1 to 3.
The results further prove the effectiveness of our CrAM model.
 


    

\begin{figure}[H]  
\setlength{\abovecaptionskip}{0.10cm}
\centering
\includegraphics[width=0.48\textwidth]{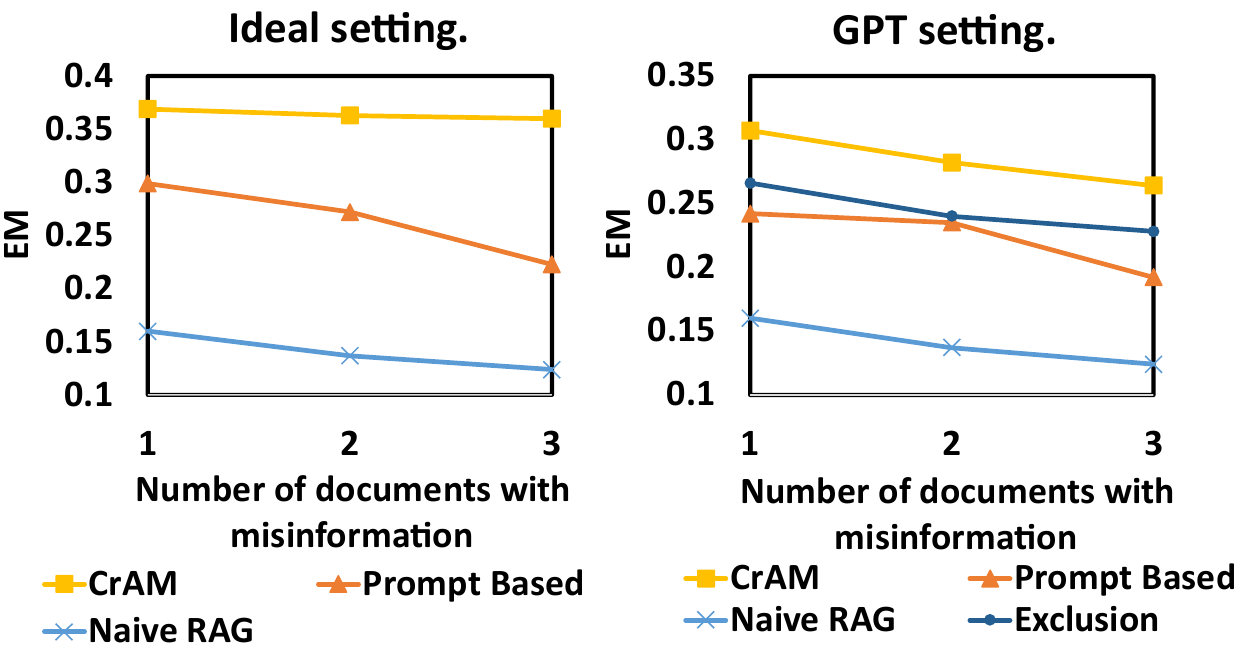}

\caption{Performance change on NQ regarding the varying number of documents with misinformation.} 
\label{fig:varing_misinformation}
\end{figure}

\subsection{In-Depth Analysis}
\paragraph{Effect of Number of Low-credibility Documents.}
In the following, we analyze the effect of varying the number of low-credibility documents fed into the LLM.
We conduct experiments using Llama3-8B on the NQ dataset.
Specifically, we vary the number of low-credibility documents from $1$ to $3$ while keeping the number of high-credibility documents constant, i.e., $4$.
We present the experimental results in Figure~\ref{fig:varing_misinformation}.
From the figure, we make the following observations. 
1) Our CrAM model consistently outperforms the compared models when changing the number of low-credibility documents from 1 to 3 in both ideal and GPT settings.
2) Comparably, our CrAM model exhibits much smaller performance drops compared to other models when increasing the number of low-credibility documents.
These results demonstrate the robustness of our proposed model to the varying number of low-credibility documents.


\begin{figure}[ht] 
\setlength{\abovecaptionskip}{0.10cm}
\setlength{\belowcaptionskip}{-0.30cm}
\centering
  \includegraphics[width=0.48\textwidth]{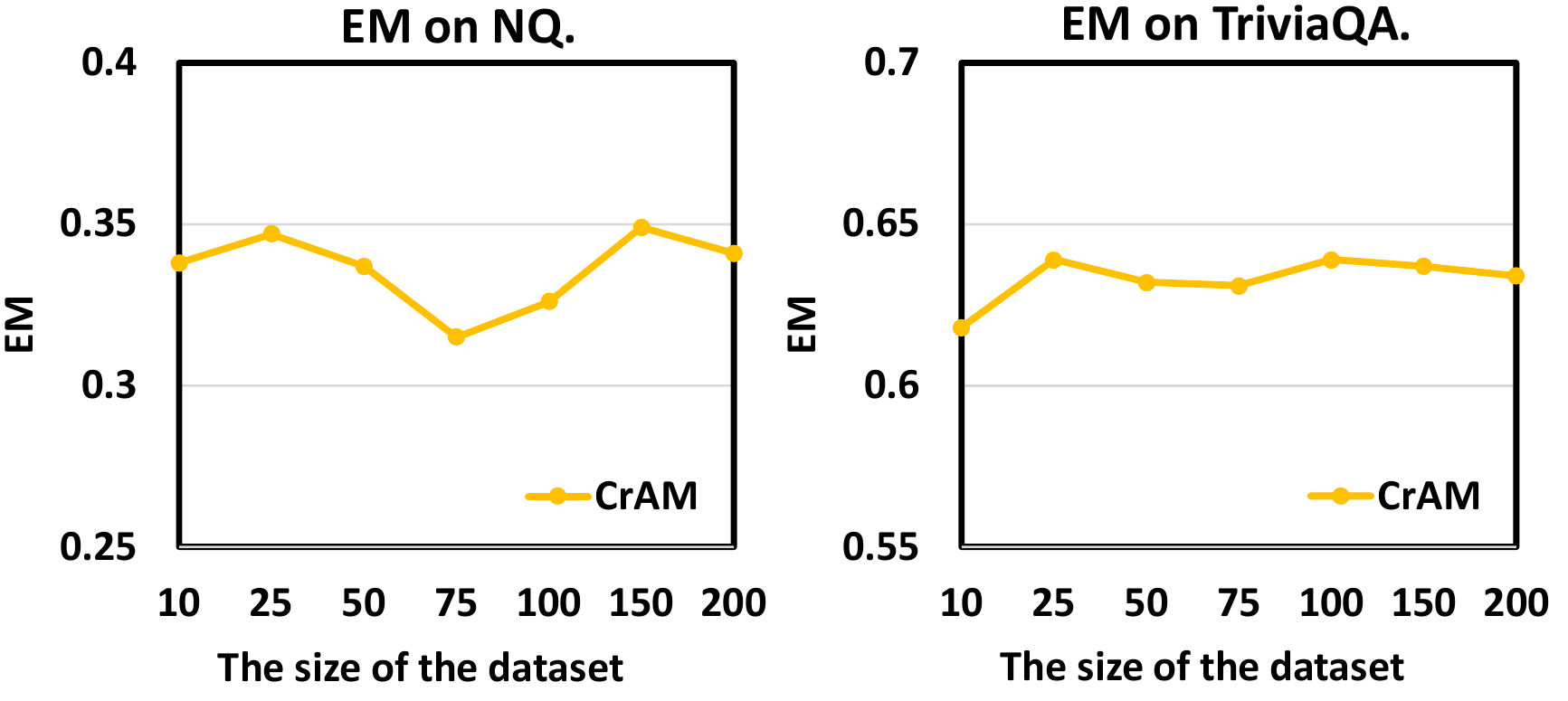}
  \caption{Performance on NQ and TriviaQA regarding the dataset size for determining the influential attention head changes.} 
  \label{fig:varing_datanum}
\end{figure}
\label{sec:data size impact}
\paragraph{Effect of Dataset Size on Attention Heads Selection.}
As we described in Section \ref{sec_cram}, we randomly select $100$ data points from each dataset to identify the influential attention heads. 
In the following, we vary the number of data points used for selecting these influential attention heads to analyze its impact on model performance.
The experimental results are presented in Figure~\ref{fig:varing_datanum}.
Despite fluctuations in performance along with the changing dataset size, the variations are not substantial on both NQ and TriviaQA datasets, with a maximum difference of $4\%$ in terms of EM. 
The results indicate that the number of data points has a minor impact on the final model performance.

\begin{figure}[t]
\setlength{\abovecaptionskip}{0.10cm}
\setlength{\belowcaptionskip}{-0.30cm}
  \centering
  \includegraphics[width=0.48\textwidth]{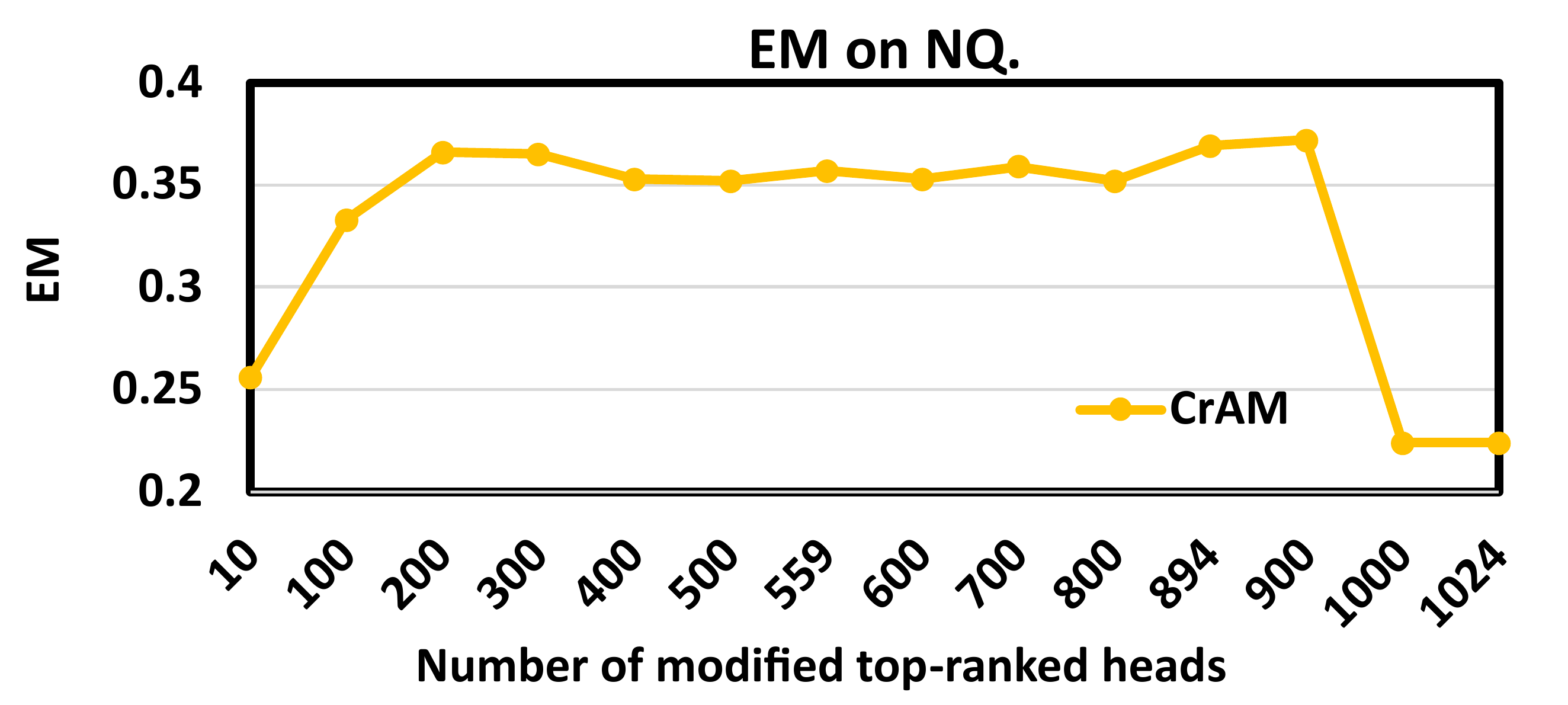}
  \caption{Performance on NQ in ideal setting regarding the varying number of selected attention heads.}
  \label{fig:different head num}
\end{figure}

\frenchspacing  
\paragraph{Analysis on Number of Selected Attention Heads.} 
In the following, we analyze the performance change when we adjust the number of selected attention heads.
We present the results in Figure~\ref{fig:different head num}.

\begin{figure}[t!]
\setlength{\abovecaptionskip}{-0.20cm}
\setlength{\belowcaptionskip}{-0.3cm}
  \centering
  \includegraphics[width=0.5\textwidth]{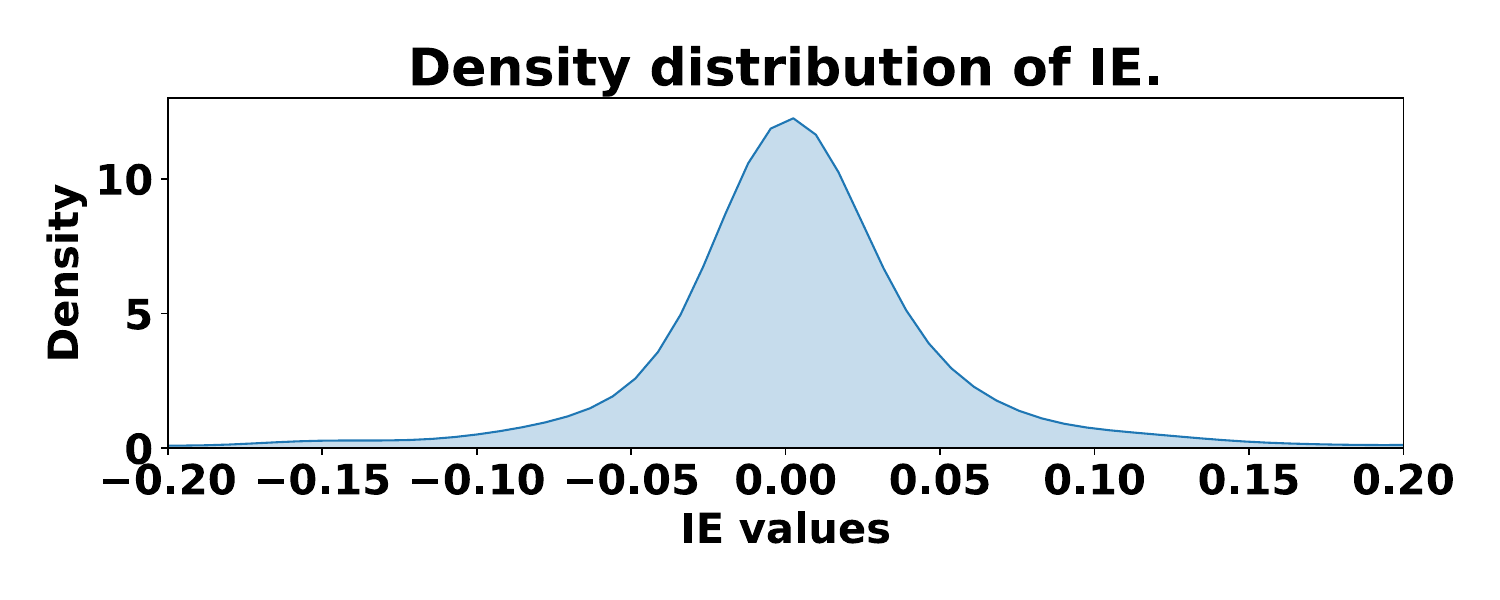}
  \caption{Density distribution of IE of all the attention heads in Llama3-8B.}
  \label{fig:dense distribution}
\end{figure}
We observe a sharp drop in model performance when the number of selected attention heads is near either 0 or the maximum number of heads, i.e., 1024;  comparably, it has a minor effect when the number of selected attention heads falls into the range of values in between.
To investigate the underlying reasons, we further analyze the IE's density distribution using Llama3-8B, as shown in Figure~\ref{fig:dense distribution}.
We find that the IE density distribution approximates a normal distribution centered around 0, with the majority of values concentrated near 0.
It indicates that most attention heads have minor impact on model performance, and only when the attention heads with IE values far from zero, either positive or negative, are selected, the model performance will be affected significantly.






\subsubsection{Ablation Study}

To better understand the rationality of our model design, we conduct ablation study and present the results in Table~\ref{tab:ablation 1+4 ideal}.
\begin{table}[t]
\centering
\setlength{\tabcolsep}{3mm}{
\begin{adjustbox}{max width=0.48\textwidth}
\begin{tabular}{llcc}
\toprule
\multirow{2}{*}{Model} & \multirow{2}{*}{Method} & \multicolumn{1}{c}{NQ} & \multicolumn{1}{c}{TriviaQA} \\
\cmidrule(lr){3-4} 
& & EM & EM \\
\midrule
\multirow{3}{*}{Qwen1.5-7B}&  CrAM & {29.10} & {52.90} \\
&  CrAM-all & {27.20} (\textcolor{red}{-1.90}) & {50.60} (\textcolor{red}{-2.30})\\
&  Naive RAG & 10.50 (\textcolor{red}{-18.60})  & 25.00 (\textcolor{red}{-27.90})  \\
\midrule
\multirow{3}{*}{Llama2-13B} & CrAM & {33.60} & {59.90}\\
&  CrAM-all & {29.50} (\textcolor{red}{-4.10}) & {59.50} (\textcolor{red}{-0.40})\\
&  Naive RAG & 11.90 (\textcolor{red}{-21.70}) & 28.00 (\textcolor{red}{-27.90})\\ 
\midrule
\multirow{3}{*}{Llama3-8B} & CrAM & {36.90} & {64.40}\\  
&  CrAM-all & {22.40} (\textcolor{red}{-14.50}) & {51.50} (\textcolor{red}{-12.90}) \\
&  Naive RAG & 16.00 (\textcolor{red}{-20.90}) & 36.80 (\textcolor{red}{-27.60})\\
\bottomrule
\end{tabular}
\end{adjustbox}
}
\caption{Results of ablation study under ideal setting with 4 \ding{51} + 1 \ding{55} (i.e., four high-credibility documents plus one low-credibility document).}
\label{tab:ablation 1+4 ideal}
\end{table}

First, we remove the selection of influential attention heads and apply attention weight modification on all attention heads in LLMs, and denote this variant model as CrAM-all. 
As shown in Table~\ref{tab:ablation 1+4 ideal}, we observe that the performance of the CrAM-all model has noticeable drops on all three LLMs. 
Among them, Llama3-8B based CrAM has the largest decrease on both NQ and TriviaQA, i.e.,  $14.5\%$ and $12.9\%$.
This indicates the necessity of identifying the influential attention heads before modifying the attention weights.

If we disable the attention weight modification mechanism in our model, it becomes the Naive RAG method.
Table~\ref{tab:ablation 1+4 ideal} shows that this results in a remarkable performance drop on all three LLMs compared to the CrAM model.
For instance, the performance of all three LLMs decreases more than $27.5\%$ on TriviaQA dataset.
These results verify that it is necessary to modify the attention weight and meanwhile take into account the credibility scores of the documents.

\section{Related Work}
\paragraph{Misinformation Detection.}
Misinformation detection aims to identify false or misleading information from various data sources \citep{guo2019future,8944587,vaibhav-etal-2019-sentence,huang2024effective}.
It can be categorized into non-LLM-based methods and LLM-based methods. 
Non-LLM methods often involve training models to identify misinformation \citep{vaibhav-etal-2019-sentence,Kaliyar2021,liu-etal-2023-interpretable,9325477}. For example, \citet{Kaliyar2021} utilize BERT \citep{devlin-etal-2019-bert} to score the credibility of documents, while \citet{vaibhav-etal-2019-sentence} use a graph neural network for misinformation detection. 
Comparably, LLM-based methods typically use LLMs without additional training \citep{pelrine-etal-2023-towards,10.3389/frai.2024.1341697,10174450,hoes_altay_bermeo_2023}. For instance, \citet{pelrine-etal-2023-towards} adopt GPT-4 \citep{openai2024gpt4} for document credibility scoring, while \citet{10.3389/frai.2024.1341697} employ an LLM agent \citep{xi2023rise} for iterative verification of document credibility. 
In this study, we employ LLMs to obtain the credibility score for each document similar to the previous LLM-based methods \citep{pelrine-etal-2023-towards,hoes_altay_bermeo_2023}.


\paragraph{Combating Misinformation in RAG.} 
Retrieval-Augmented Generation (RAG) enhance LLMs by retrieving relevant documents from external corpus \citep{NEURIPS2020_6b493230,izacard-grave-2021-leveraging,cai2024large}. 
However, prior works \citep{zou2024poisonedrag,pan-etal-2023-risk,pan-etal-2023-attacking} find that RAG is vulnerable to misinformation in its corpus, leading to undesired results. 
To combat misinformation in RAG, lots of studies have been conducted. 
For example, CAR~\cite{weller-etal-2024-defending} adopt a query augmentation scheme to retrieve a larger set of documents first and then apply a voting mechanism to mitigate the impact of misinformation. 
RobustRAG~\cite{xiang2024certifiably} obtains the LLM response for each document independently and aggregates these responses through keyword-based and decoding-based algorithms to generate the final result.
\citet{hong2024whyso} and \citet{pan2024notall} assign each retrieved document a credibility score and fine-tune LLMs with the documents and their scores, enabling the LLMs to leverage these credibility scores when generating.
CD$^2$~\citet{jin-etal-2024-tug-war} train two LLMs to generate truthful answers and misleading answers respectively to make it better distinguish the conflict information.
However, CAR~\cite{weller-etal-2024-defending} and RobustRAG~\cite{xiang2024certifiably} require multiple rounds of model inference, leading to inefficiency.
The methods proposed by \citet{hong2024whyso}, \citet{pan2024notall}, and \citet{jin-etal-2024-tug-war} require fine-tuning LLMs, which demands additional computational resources and well-designed training data, thereby limiting their application scenarios.

\section{Conclusion}
This work introduces CrAM, a plug-and-play method that enables RAG to automatically adjust the influence of retrieved documents on the output of LLMs based on document credibility. 
CrAM first identifies influential attention heads and then adjusts the attention weights of identified attention heads according to the credibility score of documents, regulating LLMs to pay less attention to the low-credibility documents. 
Empirical experiments demonstrate that, compared to vanilla RAG, CrAM improves EM performance by over 20\% on two datasets and even outperforms the baseline with SFT, demonstrating CrAM's efficiency. 
\section*{Acknowledgements}
This work is supported by the National Natural Science Foundation of China (62272437).

\bibliography{aaai25}

\begin{thebibliography}{46}
\providecommand{\natexlab}[1]{#1}

\bibitem[{Bai et~al.(2023)Bai, Bai, Chu, Cui, Dang, Deng, Fan et~al.}]{bai2023qwen}
Bai, J.; Bai, S.; Chu, Y.; Cui, Z.; Dang, K.; Deng, X.; Fan, Y.; et~al. 2023.
\newblock Qwen Technical Report.
\newblock arXiv:2309.16609.

\bibitem[{Cai et~al.(2024)Cai, Li, Wang, Zhu, Shen, Li, and Chua}]{cai2024large}
Cai, H.; Li, Y.; Wang, W.; Zhu, F.; Shen, X.; Li, W.; and Chua, T.-S. 2024.
\newblock Large Language Models Empowered Personalized Web Agents.
\newblock arXiv:2410.17236.

\bibitem[{Caramancion(2023)}]{10174450}
Caramancion, K.~M. 2023.
\newblock Harnessing the Power of ChatGPT to Decimate Mis/Disinformation: Using ChatGPT for Fake News Detection.
\newblock In \emph{2023 IEEE World AI IoT Congress (AIIoT)}, 0042--0046.

\bibitem[{Chen and Shu(2024)}]{chen2024llmgenerated}
Chen, C.; and Shu, K. 2024.
\newblock Can LLM-Generated Misinformation Be Detected?
\newblock arXiv:2309.13788.

\bibitem[{Chen et~al.(2017)Chen, Fisch, Weston, and Bordes}]{chen-etal-2017-reading}
Chen, D.; Fisch, A.; Weston, J.; and Bordes, A. 2017.
\newblock Reading {W}ikipedia to Answer Open-Domain Questions.
\newblock In Barzilay, R.; and Kan, M.-Y., eds., \emph{Proceedings of the 55th Annual Meeting of the Association for Computational Linguistics (Volume 1: Long Papers)}, 1870--1879. Vancouver, Canada: Association for Computational Linguistics.

\bibitem[{Clark et~al.(2019)Clark, Khandelwal, Levy, and Manning}]{clark-etal-2019-bert}
Clark, K.; Khandelwal, U.; Levy, O.; and Manning, C.~D. 2019.
\newblock What Does {BERT} Look at? An Analysis of {BERT}{'}s Attention.
\newblock In Linzen, T.; Chrupa{\l}a, G.; Belinkov, Y.; and Hupkes, D., eds., \emph{Proceedings of the 2019 ACL Workshop BlackboxNLP: Analyzing and Interpreting Neural Networks for NLP}, 276--286. Florence, Italy: Association for Computational Linguistics.

\bibitem[{Devlin et~al.(2019)Devlin, Chang, Lee, and Toutanova}]{devlin-etal-2019-bert}
Devlin, J.; Chang, M.-W.; Lee, K.; and Toutanova, K. 2019.
\newblock {BERT}: Pre-training of Deep Bidirectional Transformers for Language Understanding.
\newblock In Burstein, J.; Doran, C.; and Solorio, T., eds., \emph{Proceedings of the 2019 Conference of the North {A}merican Chapter of the Association for Computational Linguistics: Human Language Technologies, Volume 1 (Long and Short Papers)}, 4171--4186. Minneapolis, Minnesota: Association for Computational Linguistics.

\bibitem[{Dufour et~al.(2024)Dufour, Pathak, Samangouei, Hariri, Deshetti et~al.}]{dufour2024mis}
Dufour, N.; Pathak, A.; Samangouei, P.; Hariri, N.; Deshetti, S.; et~al. 2024.
\newblock AMMeBa: A Large-Scale Survey and Dataset of Media-Based Misinformation In-The-Wild.
\newblock arXiv:2405.11697.

\bibitem[{Elhage et~al.(2021)Elhage, Nanda, Olsson, Henighan, Joseph et~al.}]{elhage2021mathematical}
Elhage, N.; Nanda, N.; Olsson, C.; Henighan, T.; Joseph, N.; et~al. 2021.
\newblock A mathematical framework for transformer circuits.
\newblock \emph{Transformer Circuits Thread}, 1: 1.

\bibitem[{Gao et~al.(2024)Gao, Xiong, Gao, Jia, Pan, Bi, Dai, Sun, Wang, and Wang}]{gao2024retrievalaugmented}
Gao, Y.; Xiong, Y.; Gao, X.; Jia, K.; Pan, J.; Bi, Y.; Dai, Y.; Sun, J.; Wang, M.; and Wang, H. 2024.
\newblock Retrieval-Augmented Generation for Large Language Models: A Survey.
\newblock arXiv:2312.10997.

\bibitem[{Goonathilake and Kumara(2020)}]{9325477}
Goonathilake, M. D. P.~P.; and Kumara, P. P. N.~V. 2020.
\newblock CNN, RNN-LSTM Based Hybrid Approach to Detect State-of-the-Art Stance-Based Fake News on Social Media.
\newblock In \emph{2020 20th International Conference on Advances in ICT for Emerging Regions (ICTer)}, 23--28.

\bibitem[{Guo et~al.(2019)Guo, Ding, Yao, Liang, and Yu}]{guo2019future}
Guo, B.; Ding, Y.; Yao, L.; Liang, Y.; and Yu, Z. 2019.
\newblock The Future of Misinformation Detection: New Perspectives and Trends.
\newblock arXiv:1909.03654.

\bibitem[{Hoes, Altay, and Bermeo(2023)}]{hoes_altay_bermeo_2023}
Hoes, E.; Altay, S.; and Bermeo, J. 2023.
\newblock Leveraging ChatGPT for Efficient Fact-Checking.

\bibitem[{Hong et~al.(2024)Hong, Kim, Kang, Myaeng, and Whang}]{hong2024whyso}
Hong, G.; Kim, J.; Kang, J.; Myaeng, S.-H.; and Whang, J.~J. 2024.
\newblock Why So Gullible? Enhancing the Robustness of Retrieval-Augmented Models against Counterfactual Noise.
\newblock arXiv:2305.01579.

\bibitem[{Huang et~al.(2024)Huang, Zhu, Tang, Zhou, Lei, Lv, and Chua}]{huang2024effective}
Huang, Y.; Zhu, F.; Tang, J.; Zhou, P.; Lei, W.; Lv, J.; and Chua, T.-S. 2024.
\newblock Effective and Efficient Adversarial Detection for Vision-Language Models via A Single Vector.
\newblock arXiv:2410.22888.

\bibitem[{Izacard and Grave(2021)}]{izacard-grave-2021-leveraging}
Izacard, G.; and Grave, E. 2021.
\newblock Leveraging Passage Retrieval with Generative Models for Open Domain Question Answering.
\newblock In \emph{Proceedings of the 16th Conference of the European Chapter of the Association for Computational Linguistics: Main Volume}, 874--880. Online: Association for Computational Linguistics.

\bibitem[{Jin et~al.(2024)Jin, Cao, Chen, Liu, Jiang, Xu, Qiuxia, and Zhao}]{jin-etal-2024-tug-war}
Jin, Z.; Cao, P.; Chen, Y.; Liu, K.; Jiang, X.; Xu, J.; Qiuxia, L.; and Zhao, J. 2024.
\newblock Tug-of-War between Knowledge: Exploring and Resolving Knowledge Conflicts in Retrieval-Augmented Language Models.
\newblock In Calzolari, N.; Kan, M.-Y.; Hoste, V.; Lenci, A.; Sakti, S.; and Xue, N., eds., \emph{Proceedings of the 2024 Joint International Conference on Computational Linguistics, Language Resources and Evaluation (LREC-COLING 2024)}, 16867--16878. Torino, Italia: ELRA and ICCL.

\bibitem[{Joshi et~al.(2017)Joshi, Choi, Weld, and Zettlemoyer}]{joshi-etal-2017-triviaqa}
Joshi, M.; Choi, E.; Weld, D.; and Zettlemoyer, L. 2017.
\newblock {T}rivia{QA}: A Large Scale Distantly Supervised Challenge Dataset for Reading Comprehension.
\newblock In Barzilay, R.; and Kan, M.-Y., eds., \emph{Proceedings of the 55th Annual Meeting of the Association for Computational Linguistics (Volume 1: Long Papers)}, 1601--1611. Vancouver, Canada: Association for Computational Linguistics.

\bibitem[{Kaliyar, Goswami, and Narang(2021)}]{Kaliyar2021}
Kaliyar, R.~K.; Goswami, A.; and Narang, P. 2021.
\newblock FakeBERT: Fake news detection in social media with a BERT-based deep learning approach.
\newblock \emph{Multimedia Tools and Applications}, 80(8): 11765--11788.

\bibitem[{Kaliyar and Singh(2019)}]{8944587}
Kaliyar, R.~K.; and Singh, N. 2019.
\newblock Misinformation Detection on Online Social Media-A Survey.
\newblock In \emph{2019 10th International Conference on Computing, Communication and Networking Technologies (ICCCNT)}, 1--6.

\bibitem[{Karpukhin et~al.(2020)Karpukhin, Oguz, Min, Lewis, Wu, Edunov, Chen, and Yih}]{karpukhin-etal-2020-dense}
Karpukhin, V.; Oguz, B.; Min, S.; Lewis, P.; Wu, L.; Edunov, S.; Chen, D.; and Yih, W.-t. 2020.
\newblock Dense Passage Retrieval for Open-Domain Question Answering.
\newblock In Webber, B.; Cohn, T.; He, Y.; and Liu, Y., eds., \emph{Proceedings of the 2020 Conference on Empirical Methods in Natural Language Processing (EMNLP)}, 6769--6781. Online: Association for Computational Linguistics.

\bibitem[{Kwiatkowski et~al.(2019)Kwiatkowski, Palomaki, Redfield, Collins et~al.}]{nq}
Kwiatkowski, T.; Palomaki, J.; Redfield, O.; Collins, M.; et~al. 2019.
\newblock Natural Questions: A Benchmark for Question Answering Research.
\newblock \emph{Transactions of the Association for Computational Linguistics}, 7: 453--466.

\bibitem[{Lewis et~al.(2020)Lewis, Perez, Piktus, Petroni, Karpukhin, Goyal, K\"{u}ttler, Lewis, Yih, Rockt\"{a}schel, Riedel, and Kiela}]{NEURIPS2020_6b493230}
Lewis, P.; Perez, E.; Piktus, A.; Petroni, F.; Karpukhin, V.; Goyal, N.; K\"{u}ttler, H.; Lewis, M.; Yih, W.-t.; Rockt\"{a}schel, T.; Riedel, S.; and Kiela, D. 2020.
\newblock Retrieval-Augmented Generation for Knowledge-Intensive NLP Tasks.
\newblock In Larochelle, H.; Ranzato, M.; Hadsell, R.; Balcan, M.; and Lin, H., eds., \emph{Advances in Neural Information Processing Systems}, volume~33, 9459--9474. Curran Associates, Inc.

\bibitem[{Li et~al.(2024)Li, Wang, Feng, Zhu, Wang, and Chua}]{li2024think}
Li, M.; Wang, W.; Feng, F.; Zhu, F.; Wang, Q.; and Chua, T.-S. 2024.
\newblock Think Twice Before Trusting: Self-Detection for Large Language Models through Comprehensive Answer Reflection.
\newblock In Al-Onaizan, Y.; Bansal, M.; and Chen, Y.-N., eds., \emph{Findings of the Association for Computational Linguistics: EMNLP 2024}, 11858--11875. Association for Computational Linguistics.

\bibitem[{Liu, Wang, and Li(2023)}]{liu-etal-2023-interpretable}
Liu, H.; Wang, W.; and Li, H. 2023.
\newblock Interpretable Multimodal Misinformation Detection with Logic Reasoning.
\newblock In Rogers, A.; Boyd-Graber, J.; and Okazaki, N., eds., \emph{Findings of the Association for Computational Linguistics: ACL 2023}, 9781--9796. Toronto, Canada: Association for Computational Linguistics.

\bibitem[{Meng et~al.(2022)Meng, Bau, Andonian, and Belinkov}]{NEURIPS2022_causal}
Meng, K.; Bau, D.; Andonian, A.; and Belinkov, Y. 2022.
\newblock Locating and Editing Factual Associations in GPT.
\newblock In Koyejo, S.; Mohamed, S.; Agarwal, A.; Belgrave, D.; Cho, K.; and Oh, A., eds., \emph{Advances in Neural Information Processing Systems}, volume~35, 17359--17372. Curran Associates, Inc.

\bibitem[{Meta(2024)}]{llama3}
Meta. 2024.
\newblock LlaMA 3.

\bibitem[{OpenAI et~al.(2024)OpenAI, Achiam, Adler, Agarwal, Ahmad et~al.}]{openai2024gpt4}
OpenAI; Achiam, J.; Adler, S.; Agarwal, S.; Ahmad, L.; et~al. 2024.
\newblock GPT-4 Technical Report.
\newblock arXiv:2303.08774.

\bibitem[{Pan et~al.(2023{\natexlab{a}})Pan, Chen, Kan, and Wang}]{pan-etal-2023-attacking}
Pan, L.; Chen, W.; Kan, M.-Y.; and Wang, W.~Y. 2023{\natexlab{a}}.
\newblock Attacking Open-domain Question Answering by Injecting Misinformation.
\newblock In Park, J.~C.; Arase, Y.; Hu, B.; Lu, W.; Wijaya, D.; Purwarianti, A.; and Krisnadhi, A.~A., eds., \emph{Proceedings of the 13th International Joint Conference on Natural Language Processing and the 3rd Conference of the Asia-Pacific Chapter of the Association for Computational Linguistics (Volume 1: Long Papers)}, 525--539. Nusa Dua, Bali: Association for Computational Linguistics.

\bibitem[{Pan et~al.(2024)Pan, Cao, Lin, Han, Zheng, Wang, Cai, and Sun}]{pan2024notall}
Pan, R.; Cao, B.; Lin, H.; Han, X.; Zheng, J.; Wang, S.; Cai, X.; and Sun, L. 2024.
\newblock Not All Contexts Are Equal: Teaching LLMs Credibility-aware Generation.
\newblock arXiv:2404.06809.

\bibitem[{Pan et~al.(2023{\natexlab{b}})Pan, Pan, Chen, Nakov, Kan, and Wang}]{pan-etal-2023-risk}
Pan, Y.; Pan, L.; Chen, W.; Nakov, P.; Kan, M.-Y.; and Wang, W. 2023{\natexlab{b}}.
\newblock On the Risk of Misinformation Pollution with Large Language Models.
\newblock In Bouamor, H.; Pino, J.; and Bali, K., eds., \emph{Findings of the Association for Computational Linguistics: EMNLP 2023}, 1389--1403. Singapore: Association for Computational Linguistics.

\bibitem[{Pelrine et~al.(2023)Pelrine, Imouza, Thibault, Reksoprodjo, Gupta, Christoph, Godbout, and Rabbany}]{pelrine-etal-2023-towards}
Pelrine, K.; Imouza, A.; Thibault, C.; Reksoprodjo, M.; Gupta, C.; Christoph, J.; Godbout, J.-F.; and Rabbany, R. 2023.
\newblock Towards Reliable Misinformation Mitigation: Generalization, Uncertainty, and {GPT}-4.
\newblock In Bouamor, H.; Pino, J.; and Bali, K., eds., \emph{Proceedings of the 2023 Conference on Empirical Methods in Natural Language Processing}, 6399--6429. Singapore: Association for Computational Linguistics.

\bibitem[{Quelle and Bovet(2024)}]{10.3389/frai.2024.1341697}
Quelle, D.; and Bovet, A. 2024.
\newblock The perils and promises of fact-checking with large language models.
\newblock \emph{Frontiers in Artificial Intelligence}, 7.

\bibitem[{Rajpurkar et~al.(2016)Rajpurkar, Zhang, Lopyrev, and Liang}]{rajpurkar-etal-2016-squad}
Rajpurkar, P.; Zhang, J.; Lopyrev, K.; and Liang, P. 2016.
\newblock {SQ}u{AD}: 100,000+ Questions for Machine Comprehension of Text.
\newblock In Su, J.; Duh, K.; and Carreras, X., eds., \emph{Proceedings of the 2016 Conference on Empirical Methods in Natural Language Processing}, 2383--2392. Austin, Texas: Association for Computational Linguistics.

\bibitem[{Touvron et~al.(2023)Touvron, Martin, Stone, Albert, Almahairi et~al.}]{touvron2023llama}
Touvron, H.; Martin, L.; Stone, K.; Albert, P.; Almahairi, A.; et~al. 2023.
\newblock Llama 2: Open Foundation and Fine-Tuned Chat Models.
\newblock arXiv:2307.09288.

\bibitem[{Vaibhav, Mandyam, and Hovy(2019)}]{vaibhav-etal-2019-sentence}
Vaibhav, V.; Mandyam, R.; and Hovy, E. 2019.
\newblock Do Sentence Interactions Matter? Leveraging Sentence Level Representations for Fake News Classification.
\newblock In Ustalov, D.; Somasundaran, S.; Jansen, P.; Glava{\v{s}}, G.; Riedl, M.; Surdeanu, M.; and Vazirgiannis, M., eds., \emph{Proceedings of the Thirteenth Workshop on Graph-Based Methods for Natural Language Processing (TextGraphs-13)}, 134--139. Hong Kong: Association for Computational Linguistics.

\bibitem[{Vaswani et~al.(2017)Vaswani, Shazeer, Parmar, Uszkoreit, Jones, Gomez, Kaiser, and Polosukhin}]{NIPS2017_transformers}
Vaswani, A.; Shazeer, N.; Parmar, N.; Uszkoreit, J.; Jones, L.; Gomez, A.~N.; Kaiser, L.~u.; and Polosukhin, I. 2017.
\newblock Attention is All you Need.
\newblock In Guyon, I.; Luxburg, U.~V.; Bengio, S.; Wallach, H.; Fergus, R.; Vishwanathan, S.; and Garnett, R., eds., \emph{Advances in Neural Information Processing Systems}, volume~30. Curran Associates, Inc.

\bibitem[{Vincent(2023)}]{vincent2023google}
Vincent, J. 2023.
\newblock Google and Microsoft’s chatbots are already citing one another’s misinformation.
\newblock \emph{The Verge}.
\newblock Accessed: 2023-06-05.

\bibitem[{Voita et~al.(2019)Voita, Talbot, Moiseev, Sennrich, and Titov}]{voita-etal-2019-analyzing}
Voita, E.; Talbot, D.; Moiseev, F.; Sennrich, R.; and Titov, I. 2019.
\newblock Analyzing Multi-Head Self-Attention: Specialized Heads Do the Heavy Lifting, the Rest Can Be Pruned.
\newblock In Korhonen, A.; Traum, D.; and M{\`a}rquez, L., eds., \emph{Proceedings of the 57th Annual Meeting of the Association for Computational Linguistics}, 5797--5808. Florence, Italy: Association for Computational Linguistics.

\bibitem[{Weller et~al.(2024)Weller, Khan, Weir, Lawrie, and Van~Durme}]{weller-etal-2024-defending}
Weller, O.; Khan, A.; Weir, N.; Lawrie, D.; and Van~Durme, B. 2024.
\newblock Defending Against Disinformation Attacks in Open-Domain Question Answering.
\newblock In Graham, Y.; and Purver, M., eds., \emph{Proceedings of the 18th Conference of the European Chapter of the Association for Computational Linguistics (Volume 2: Short Papers)}, 402--417. St. Julian{'}s, Malta: Association for Computational Linguistics.

\bibitem[{Xi et~al.(2023)Xi, Chen, Guo, He, Ding, Hong et~al.}]{xi2023rise}
Xi, Z.; Chen, W.; Guo, X.; He, W.; Ding, Y.; Hong, B.; et~al. 2023.
\newblock The Rise and Potential of Large Language Model Based Agents: A Survey.
\newblock arXiv:2309.07864.

\bibitem[{Xiang et~al.(2024)Xiang, Wu, Zhong, Wagner, Chen, and Mittal}]{xiang2024certifiably}
Xiang, C.; Wu, T.; Zhong, Z.; Wagner, D.; Chen, D.; and Mittal, P. 2024.
\newblock Certifiably Robust RAG against Retrieval Corruption.
\newblock arXiv:2405.15556.

\bibitem[{Yoran et~al.(2024)Yoran, Wolfson, Ram, and Berant}]{yoran2024making}
Yoran, O.; Wolfson, T.; Ram, O.; and Berant, J. 2024.
\newblock Making Retrieval-Augmented Language Models Robust to Irrelevant Context.
\newblock arXiv:2310.01558.

\bibitem[{Zhang et~al.(2023)Zhang, Li, Cui, Cai, Liu, Fu et~al.}]{zhang2023hallucinationSuvey}
Zhang, Y.; Li, Y.; Cui, L.; Cai, D.; Liu, L.; Fu, T.; et~al. 2023.
\newblock Siren's Song in the AI Ocean: A Survey on Hallucination in Large Language Models.
\newblock arXiv:2309.01219.

\bibitem[{Zhu et~al.(2021)Zhu, Lei, Wang, Zheng, Poria, and Chua}]{zhu2021retrieving}
Zhu, F.; Lei, W.; Wang, C.; Zheng, J.; Poria, S.; and Chua, T.-S. 2021.
\newblock Retrieving and Reading: A Comprehensive Survey on Open-domain Question Answering.
\newblock arXiv:2101.00774.

\bibitem[{Zou et~al.(2024)Zou, Geng, Wang, and Jia}]{zou2024poisonedrag}
Zou, W.; Geng, R.; Wang, B.; and Jia, J. 2024.
\newblock PoisonedRAG: Knowledge Poisoning Attacks to Retrieval-Augmented Generation of Large Language Models.
\newblock arXiv:2402.07867.

\end{thebibliography}
\appendix
\clearpage
\section{Multi-Head Attention}
\label{sec:mha}
Currently, leading LLMs are built on autoregressive transformer architectures \citep{touvron2023llama,llama3,bai2023qwen}. 
The \textbf{multi-head attention} mechanism \citep{NIPS2017_transformers} is the core component of autoregressive transformer models. 
It is illustrated in the following steps. 

\textbf{Linear Transformation:} Given an input hidden state $\mathbf{X} \in \mathbb{R}^{n \times d}$, three linear transformations are applied to produce queries $\mathbf{Q} \in \mathbb{R}^{n \times d_k}$, keys $\mathbf{K} \in \mathbb{R}^{n \times d_k}$, and values $\mathbf{V} \in \mathbb{R}^{n \times d_v}$:
\begin{equation*}
\mathbf{Q} = \mathbf{X}\mathbf{W}^Q, \quad \mathbf{K} = \mathbf{X}\mathbf{W}^K, \quad \mathbf{V} = \mathbf{X}\mathbf{W}^V
\end{equation*}
where $\mathbf{W}^Q \in \mathbb{R}^{d \times d_k}$, $\mathbf{W}^K \in \mathbb{R}^{d \times d_k}$, and $\mathbf{W}^V \in \mathbb{R}^{d \times d_v}$ are weight matrices.

\textbf{Scaled Dot-Product Attention:} The \textbf{attention weights} are computed using the dot product of the queries and keys, scaled by $1/\sqrt{d_k}$:


\begin{equation}
\label{eq:attention score}
\mathbf{A} = \mathrm{softmax}\left(\frac{\mathbf{Q}\mathbf{K}^T}{\sqrt{d_k}}\right)
\end{equation}
The softmax function ensures that the attention weights sum to one.

\textbf{Multi-Head Attention:} Instead of performing a single attention function, $h$ attention functions (or heads) are performed in parallel. Each head has its own set of weight matrices $\mathbf{W}_i^Q, \mathbf{W}_i^K, \mathbf{W}_i^V$ and attention weights $\mathbf{A}_i$ for $i \in [1, h]$:
\begin{equation*}
\resizebox{\linewidth}{!}{$\mathrm{MultiHead}(\mathbf{Q}, \mathbf{K}, \mathbf{V}) = \mathrm{Concat}(\mathrm{head}_1, \ldots, \mathrm{head}_h)\mathbf{W}^O$}\\
\end{equation*}
where $\mathrm{head}_i = \mathbf{A}_i \mathbf{V}_i$ and $\mathbf{W}^O \in \mathbb{R}^{hd_v \times d}$ is the output weight matrix.


\section{Implementation Details}
We used \textit{gpt-3.5-turbo-0125} for all generations involving GPT, with a temperature setting of 1 to allow for variability. For Llama2-13B, Qwen1.5-7B, and Llama3-8B, we did not perform any sampling during generation to avoid randomness. For the NQ and TriviaQA datasets, we randomly selected 1,000 samples from the original test set for our evaluation. We defined \( x \) as the number of heads where \( IE > 0 \). To determine how many top-ranked heads should be included in the final modified set based on a validation set of 100 data points, we experimented with including the following quantities of heads as modification targets: \( \{1.0 \times x, 1, 1.0 \times x, 1.2 \times x, \ldots, 2.0 \times x\} \). Ultimately, we selected the hyperparameters that yielded the best EM and F1 scores on the validation set.

\begin{figure}[t]
  \centering
  \includegraphics[width=0.48\textwidth]{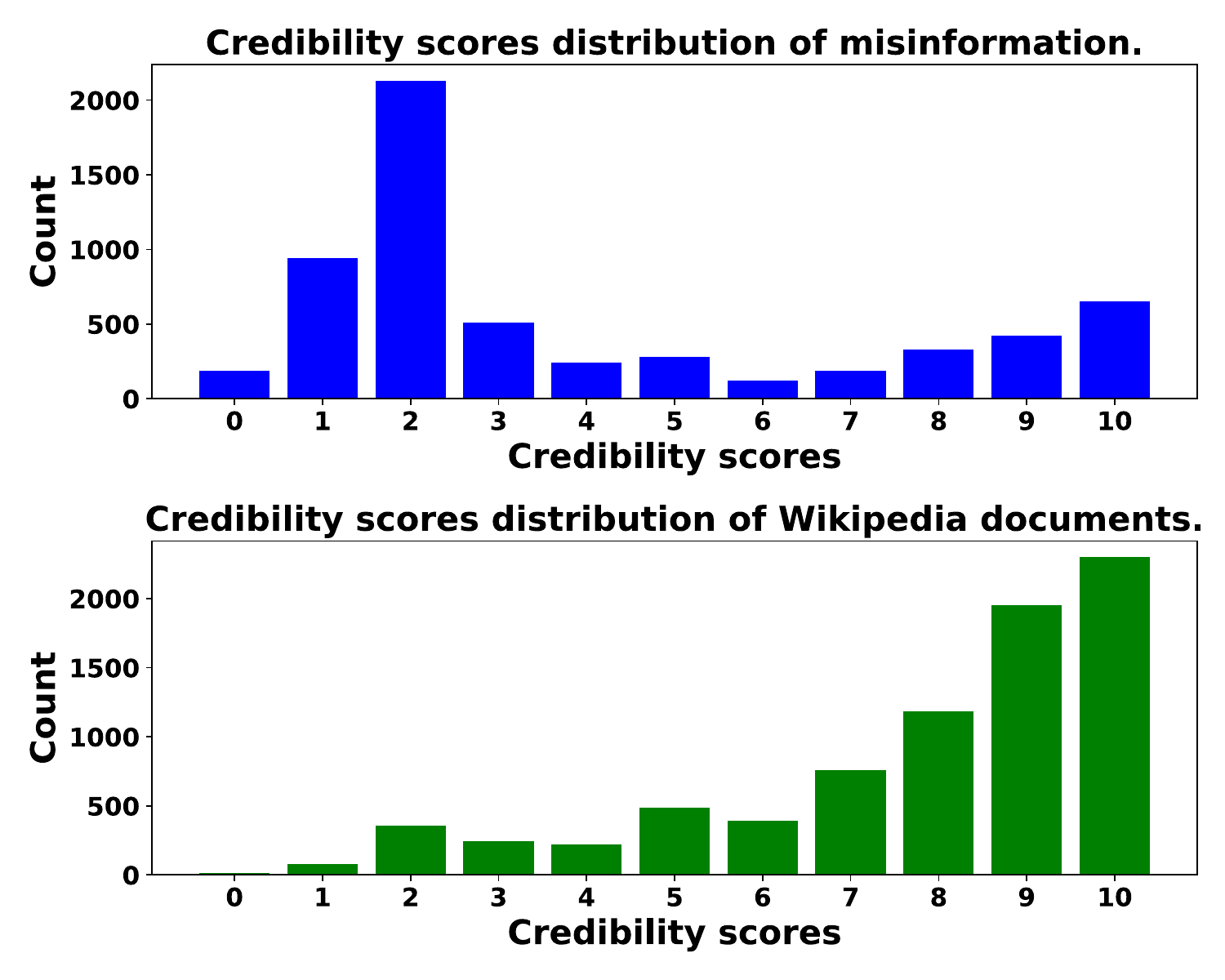}
  \caption{Distribution of GPT-generated credibility scores on misinformation and Wikipedia documents.}
  \label{fig:gpt_score_distribution}
\end{figure}

\begin{figure}[t]
  \centering
  \includegraphics[width=0.45\textwidth]{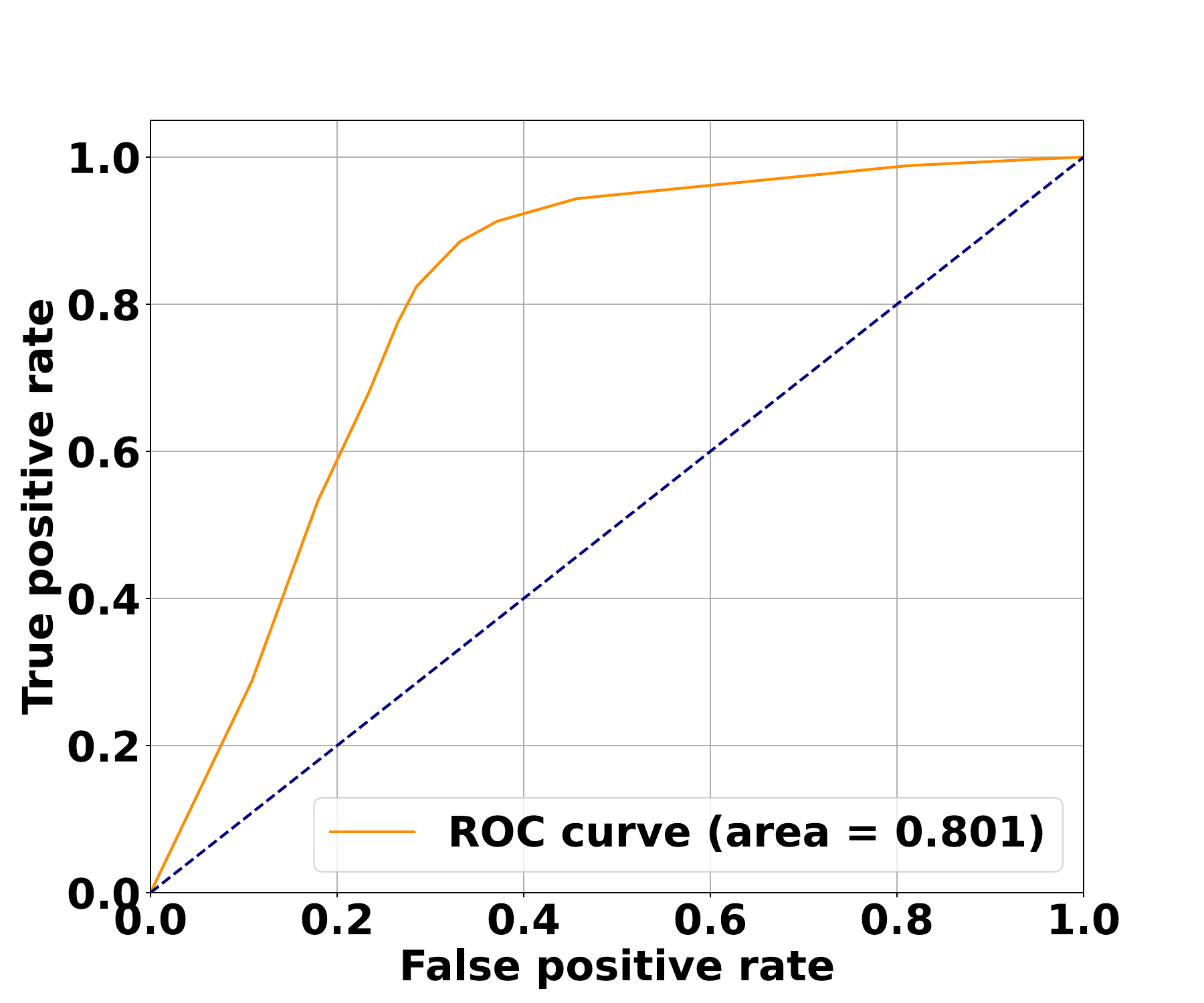}
  \caption{ROC curve of GPT-generated credibility scores, with area under curve (AUC) = 0.801.}
  \label{fig:roc}
\end{figure}

\section{GPT-Generated Credibility Scores}
\label{sec:appendix_gpt_score}
We present the distribution of GPT-generated credibility scores in Figure \ref{fig:gpt_score_distribution} and the corresponding receiver operating characteristic (ROC) curve in Figure \ref{fig:roc}. To plot the ROC curve, we begin with a dataset containing documents labeled as either misinformation (GPT-generated misinformation) or non-misinformation (retrieved factual documents). Each document receives a score from GPT ranging from 0 to 10, where lower scores indicate a higher likelihood of misinformation. For each point on the ROC curve, we select a threshold between 0 and 10. Documents scoring below this threshold are classified as misinformation (negative instances), while those scoring above are classified as non-misinformation (positive instances). We compute the false positive rate and true positive rate across all documents to gain a point in the ROC curve. This process is repeated for different thresholds, and finally, all data points are connected to form the ROC curve. Then we calculate the area under curve (AUC) of the ROC curve.

\section{Full Results with Varying Number of Documents with Misinformation}
\label{sec_full_results_different_misinformation}
\begin{figure}[t]
  \centering
  \includegraphics[width=0.45\textwidth]{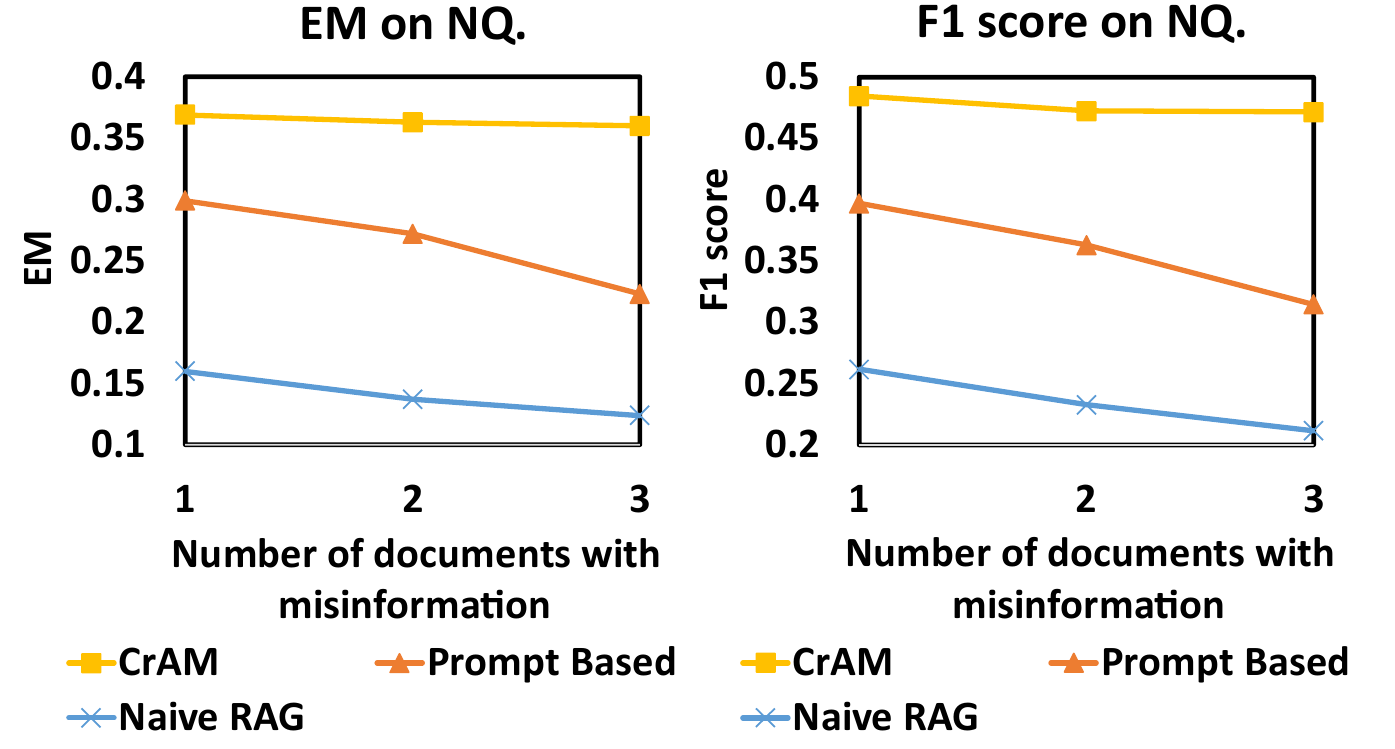}
  \caption{EM and F1 socre on NQ using Llama3-8B under ideal setting.}
  \label{fig:nq_ideal}
\end{figure}

\begin{figure}[t]
  \centering
  \includegraphics[width=0.45\textwidth]{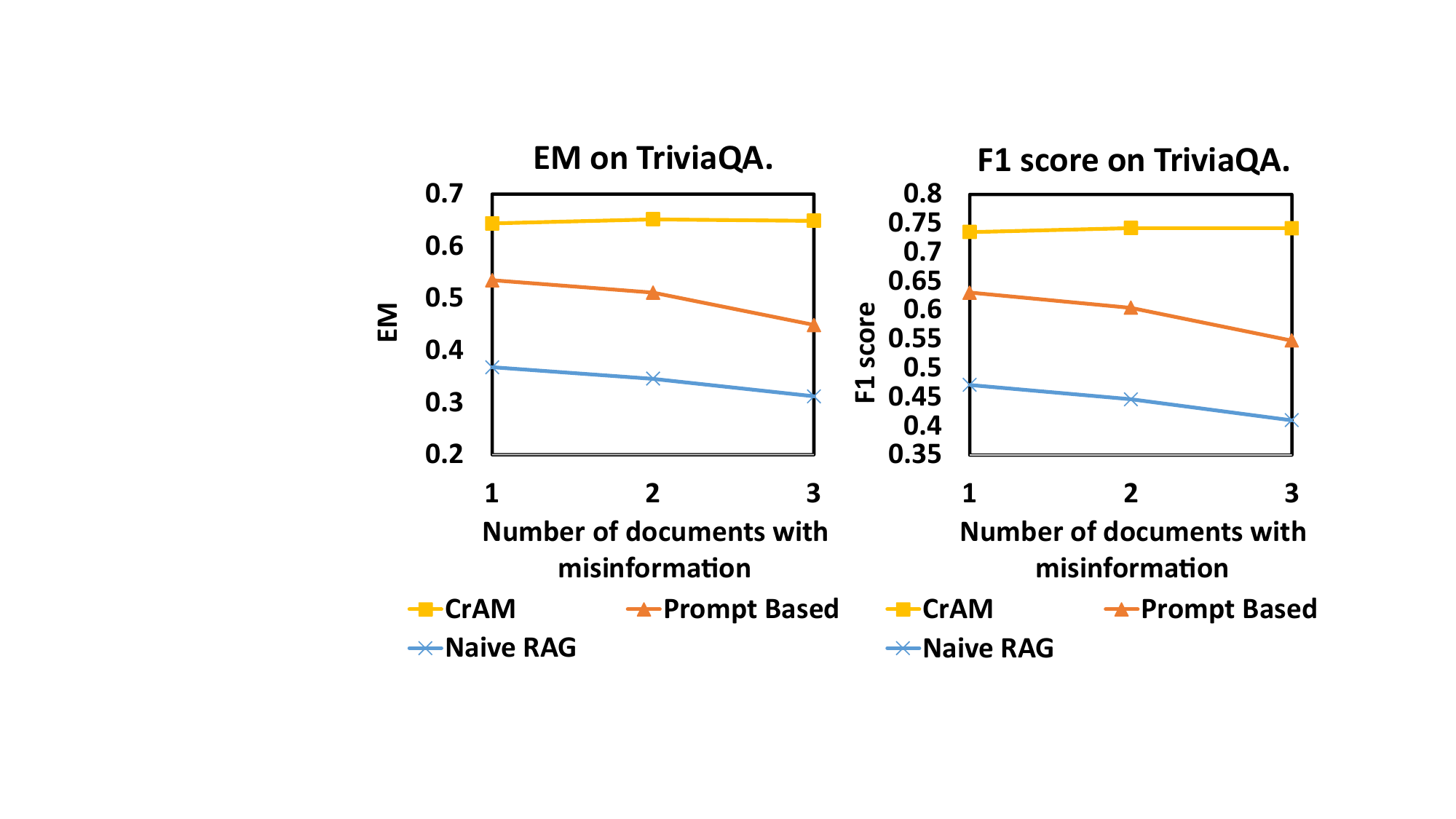}
  \caption{EM and F1 socre on TriviaQA using Llama3-8B under ideal setting.}
  \label{fig:tri_ideal}
\end{figure}

\begin{figure}[t]
  \centering
  \includegraphics[width=0.45\textwidth]{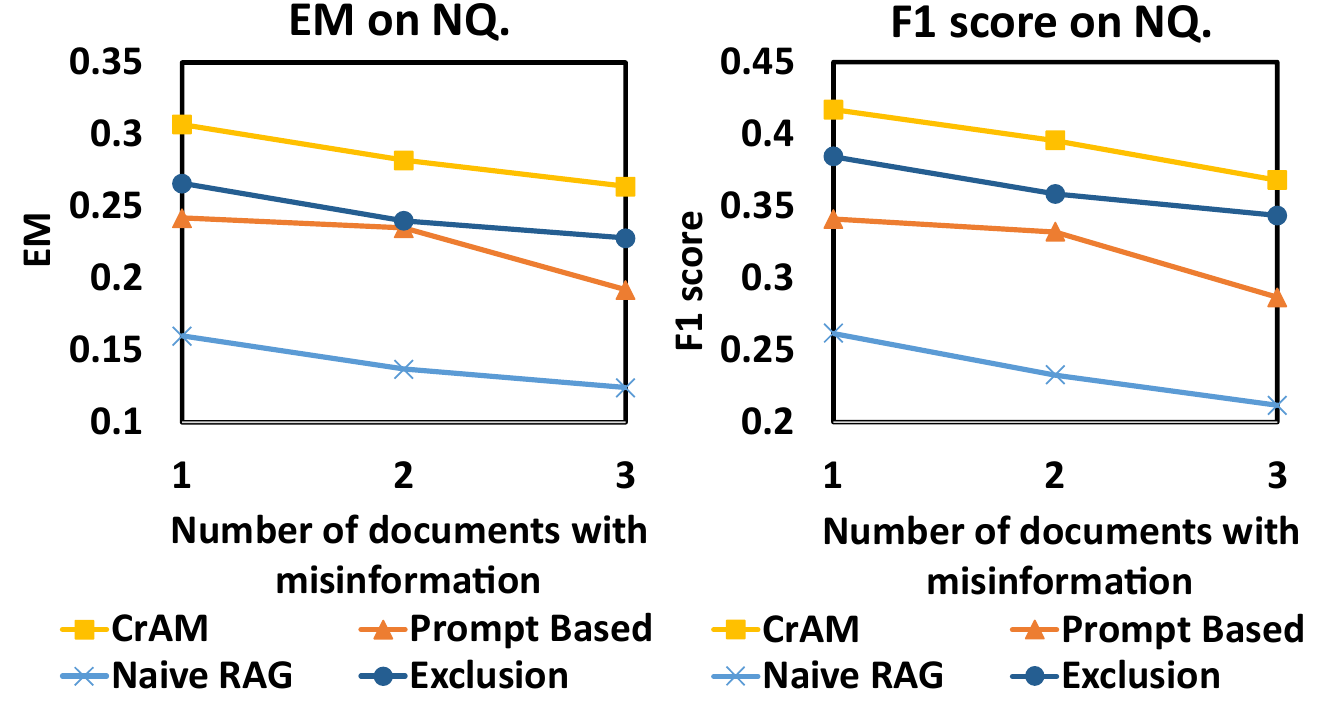}
  \caption{EM and F1 socre on NQ using Llama3-8B under GPT setting.}
  \label{fig:nq_gpt}
\end{figure}

\begin{figure}[t]
  \centering
  \includegraphics[width=0.45\textwidth]{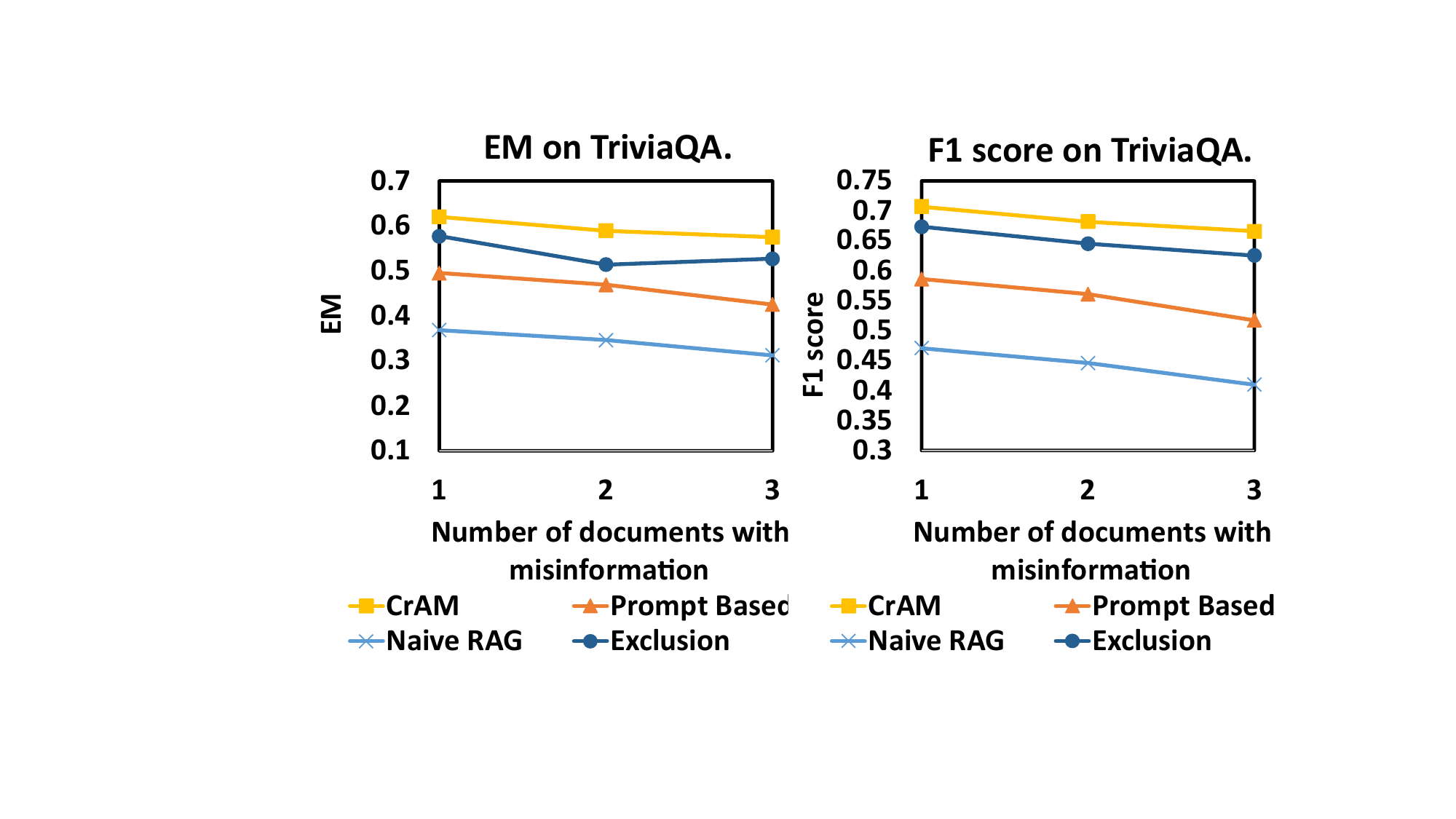}
  \caption{EM and F1 socre on TriviaQA using Llama3-8B under GPT setting.}
  \label{fig:tri_gpt}
\end{figure}

We provide the full results as the number of documents with misinformation increase, as shown in Figure \ref{fig:nq_ideal}-\ref{fig:tri_gpt}. All results are done with four correct documents.

\begin{figure}[!ht]  
\centering
\includegraphics[width=0.45\textwidth]{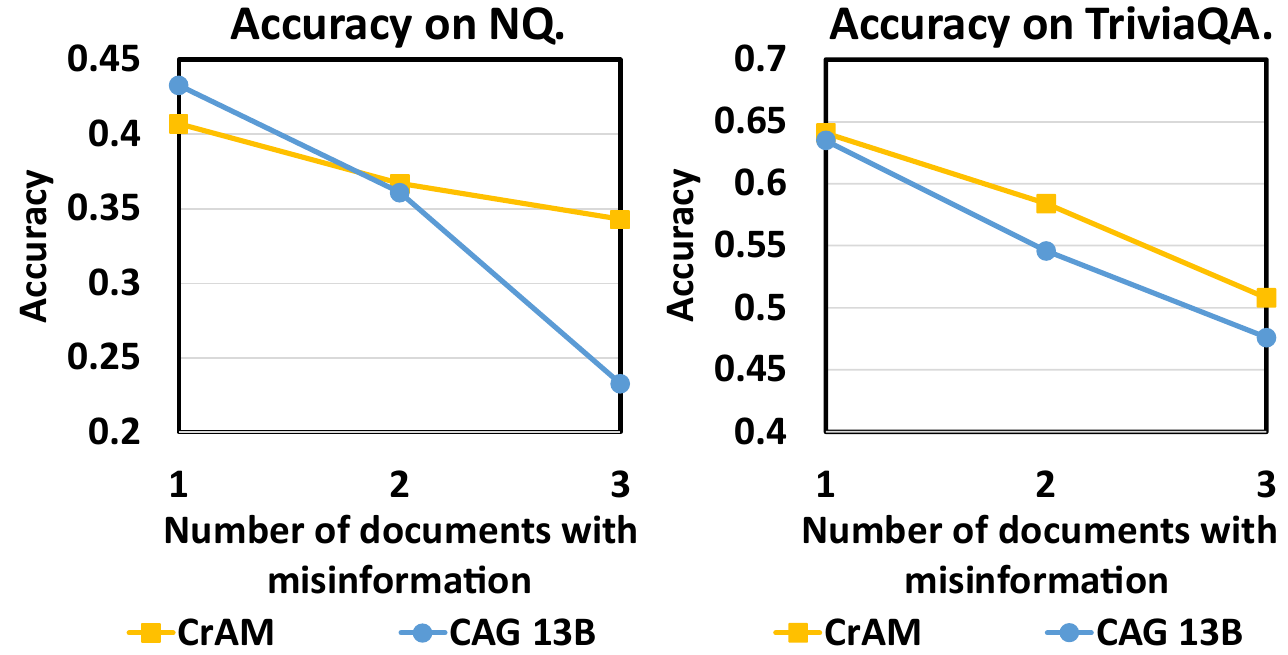}
\caption{Performance comparison of CrAM of Llama2-13B and CAG 13B with varying amounts of misinformation under ideal setting.} 
\label{fig:cag em}
\end{figure}

\section{Comparison with CAG}
\label{sec:comparison cag}
Since the CAG 13B model tends to provide lengthy responses, its performance on EM is very low. Therefore, we consider an answer "correct" if the correct answer appears in the model's prediction, and we use accuracy as the metric. The results are shown in Figure \ref{fig:cag em}. This metric is more favorable for long answers, however, CrAM still surpasses the SFT-based CAG 13B in most situations, demonstrating the superiority of our approach.

\section{Results with Filtered Misinformation}
\label{sec:results_with_filtered_misinformation}
We replaced all the correct answers in the existing misinformation with ``xxx'' (denoted as ``filtered misinformation'') and then conducted the same experiments on filtered misinformation. The results are shown in Table \ref{tab:ideal score 1+4 new}. We make the following observations. 
1) The performance of CrAM with 4 \ding{51}+ 1 \ding{55} is lower than that in Table \ref{tab:ideal score 1+4}, and it is worse than that of the Naive RAG with 4 \ding{51} in most cases. This indicates that CrAM enables LLMs to re-utilize the correct information denied by the misinformation, resulting in a better performance.
2) Table~\ref{tab:ideal score 1+4 new} demonstrates that our CrAM method still outperforms all compared methods across all three LLMs: Qwen1.5-7B, LLama2-13B, and LLama3-8B, on both NQ and TriviaQA datasets in the setting of 4 \ding{51}+ 1 \ding{55} (i.e., four high-credibility documents plus one low-credibility document), proving CrAM doesn't solely rely on correct answers in misinformation.
\begin{table*}[ht]
\centering
\setlength{\tabcolsep}{4mm}{
\begin{adjustbox}{max width=\textwidth}
\begin{tabular}{lclcccc}
\toprule
\multirow{2}{*}{Model} & \multirow{2}{*}{In-context corpus} & \multirow{2}{*}{Method} & \multicolumn{2}{c}{NQ} & \multicolumn{2}{c}{TriviaQA} \\
\cmidrule(lr){4-5} \cmidrule(lr){6-7}
& & & EM & F1 score & EM & F1 score \\
\midrule
\multirow{5}{*}{Qwen1.5-7B} & 0 \ding{51} & Naive LLM & 7.20 & 16.41 & 28.00 & 38.23 \\
& 4 \ding{51} & Naive RAG & 27.60 & 39.08 & 55.30 & 66.85 \\
\cmidrule(lr){2-7}
& \multirow{3}{*}{4 \ding{51} + 1 \ding{55}} & Naive RAG & 9.70 & 20.22 & 25.40 & 36.14 \\
&  & Prompt Based & 10.40 & 20.67 & 26.30 & 37.12 \\

&  & CrAM & \textbf{25.90} (\textcolor{red}{-1.70}) & \textbf{37.87} (\textcolor{red}{-1.21}) & \textbf{51.70} (\textcolor{red}{-3.60}) &  \textbf{63.07} (\textcolor{red}{-3.78})\\
\midrule
\multirow{5}{*}{Llama2-13B} & 0 \ding{51} & Naive LLM & 20.30 & 28.59 & 50.40 & 57.56 \\
& 4 \ding{51} & Naive RAG & 28.90 & 39.98 & 62.50 & 71.03 \\
\cmidrule(lr){2-7}
& \multirow{3}{*}{4 \ding{51} + 1 \ding{55}} & Naive RAG & 12.20 & 20.71 & 27.60 & 35.80 \\
&  & Prompt Based & 9.90 & 20.48 & 21.90 & 31.22 \\

&  & CrAM & \textbf{29.90} (\textcolor{red}{+1.00}) & \textbf{40.85} (\textcolor{red}{+0.87}) & \textbf{57.90} (\textcolor{red}{-4.60}) & \textbf{65.60} (\textcolor{red}{-5.43})\\
\midrule
\multirow{5}{*}{Llama3-8B} & 0 \ding{51} & Naive LLM & 20.60 & 30.58 & 55.70 & 62.67 \\
& 4 \ding{51} & Naive RAG & 33.10 & 45.66 & 64.30 & 73.68 \\
\cmidrule(lr){2-7}
& \multirow{3}{*}{4 \ding{51} + 1 \ding{55}} & Naive RAG & 16.10 & 26.57 & 38.70 & 48.84 \\
&  & Prompt Based & 25.20 & 35.72 & 52.10 & 61.03 \\

&  & CrAM & \textbf{33.80} (\textcolor{red}{+0.70}) & \textbf{45.63} (\textcolor{red}{-0.03}) & \textbf{63.70} (\textcolor{red}{-0.60}) & \textbf{72.87} (\textcolor{red}{-0.81}) \\
\bottomrule
\end{tabular}
\end{adjustbox}
}
\caption{With ideal credibility scores and filtered misinformation, we evaluate the performance of three models on two open-domain QA datasets. 0 \ding{51} indicates no document and the model is directly prompted, 4 \ding{51} indicates that all four documents are retrieved from the Wikipedia dump, while 4 \ding{51} + 1 \ding{55} indicates four documents from the Wikipedia dump plus one piece of misinformation. In the 4 \ding{51} + 1 \ding{55} setting, the best performance is highlighted in \textbf{bold}. And the \textcolor{red}{red} part indicates the difference between Naive RAG with 4 \ding{51} and CrAM with 4 \ding{51} + 1 \ding{55}.}
\label{tab:ideal score 1+4 new}
\end{table*}

\section{Time Analysis}
Compared to fine-tuning, our method has a significantly lower time overhead. We only need a validation set to determine the modified attention heads for the whole dataset. We will explain all the computational costs of CrAM step by step. The extra computational overhead of CrAM primarily stems from two aspects:
\begin{itemize}
    \item Modifying all the attention heads across all the transformer layers to rank the influential heads.
    \item Using a validation set to determine the number of attention heads for modification.
\end{itemize}
For the first part, on an A40 GPU, modifying all attention heads of the Llama3-8B model for a single data point takes 2 minutes. As we can see from Figure 5, just 10 data points are sufficient to achieve significant performance, so this step can be completed in 20 minutes. This step involves calculating the probabilities for certain answers and thus requires only a single inference, which is relatively fast. For the second part, we used 75 validation data points in our experiments and tested the performance of 10 different numbers of attention heads, which took 23 minutes. Therefore, the total additional time for a dataset requires less than 43 minutes.

\section{Prompts}
\label{sec:prompts}
We have made all the prompts used in our experiments publicly available, as shown in Figure \ref{fig:prompt_fake_gen}-\ref{fig:prompt_for_prompt_based}.

\begin{figure}[t]
\centering
\begin{tcolorbox}[colframe=black, colback=white, rounded corners]
Please generate a text that is logical and supports the statement: The answer to ``\{Question\}'' is ``\{Wrong answer\}'', rather than ``\{Correct answer\}''. The generated text should be in the following form: CNN news. The generated text should be less than 200 words. Just output the generated text , and do not output anything else.\\Generated Text:
\end{tcolorbox}
\caption{Prompt for misinformation generation.}
\label{fig:prompt_fake_gen}
\end{figure}

\begin{figure}[t]
\centering
\begin{tcolorbox}[colframe=black, colback=white, rounded corners]
Answer the following question based on your internal knowledge with one or few words.\\Question: \{Question\}\\Answer: 
\end{tcolorbox}
\caption{Prompt for Naive LLM without documents.}
\end{figure}

\begin{figure}[t]
\centering
\begin{tcolorbox}[colframe=black, colback=white, rounded corners]
Given the following information: \\\{Documents\}\\Answer the following question based on the given information or your internal knowledge with one or few words without the source.\\Question: \{Question\}\\Answer: 
\end{tcolorbox}
\caption{Prompt for RAG.}
\end{figure}

\begin{figure}[t]
\centering
\begin{tcolorbox}[colframe=black, colback=white, rounded corners]
You are an assistant who can answer questions based on the given passages. Each passage has a credibility score that indicates the relevance and accuracy of the passage to the question. Your answer need to combine multiple passages and their credibility. Question: \{Question\}\\Docs: \{Documents\}\\\\Your answer should based on the given information or your internal knowledge with one or few words without the source  (just output a answer, don't output anything else). Answer:
\end{tcolorbox}
\caption{Prompt for CAG.}
\end{figure}

\begin{figure}[t]
\centering
\begin{tcolorbox}[colframe=black, colback=white, rounded corners]
You are an assistant capable of answering questions based on provided passages. Each passage is assigned a truthfulness score (0-10), where a higher score indicates greater credibility. Your answer need to combine multiple passages and their credibility. Given the following information: \\\{Documents\}\\Answer the following question based on the given information or your internal knowledge with one or few words without the source (just output a answer, don't output anything else).\\Question: \{Question\}\\Answer:
\end{tcolorbox}
\caption{Prompt for prompt-based method.}
\label{fig:prompt_for_prompt_based}
\end{figure}

\begin{figure*}[t]
\centering
\begin{tcolorbox}[colframe=black, colback=white, rounded corners]
Your task is to evaluate the authenticity of a text based on your internal knowledge. Specifically, I will provide you with a passage that may contain accurate information or fabricated errors. Using your own knowledge, reason, and deduction, you are to assign a credibility score ranging from 0 to 10, where a higher score indicates greater authenticity and a lower score suggests lesser authenticity. \\Here are 2 examples (you should follow the output format below):\\\#\#\#\#\#\#\#\#\#\#\\Passage:\\In a groundbreaking discovery, researchers have found that Albert Einstein was the first recipient of the Nobel Prize in Physics. According to newly uncovered documents, Einstein's pioneering work in theoretical physics, particularly his theory of relativity, was recognized by the Nobel Committee in 1921. This revelation challenges the long-held belief that Marie Curie was the first Nobel laureate in physics, and solidifies Einstein's place as one of the greatest minds in scientific history.\\\\Analysis:\\1. Albert Einstein as the First Nobel Prize Recipient in Physics: This is incorrect. The first Nobel Prize in Physics was awarded in 1901, not to Albert Einstein, but to Wilhelm Conrad Röntgen for the discovery of X-rays.\\2. Einstein's Nobel Prize Recognition: Albert Einstein was indeed awarded the Nobel Prize in Physics in 1921, but not for his theory of relativity. He received it for his discovery of the photoelectric effect, which was instrumental in the development of quantum theory.\\3. Marie Curie as the First Nobel Laureate in Physics: This is also incorrect. Marie Curie was a Nobel laureate, but she was not the first to win the Nobel Prize in Physics. Her first Nobel Prize was in Physics in 1903, shared with her husband Pierre Curie and Henri Becquerel for their work on radioactivity. Marie Curie was, notably, the first woman to win a Nobel Prize, and the first person to win Nobel Prizes in two different scientific fields (Physics and Chemistry).\\4. Implication about the Nobel Committee's Recognition of Relativity: As mentioned, Einstein's Nobel Prize was not for relativity, despite its profound impact on physics. The Nobel Committee specifically avoided awarding the prize for relativity at the time due to ongoing debates and lack of experimental confirmation of the theory during that period.\\\\Credibility Score: 0\\\\\\Passage:\\The first Nobel Prize in Physics was awarded to Wilhelm Conrad Roentgen in 1901. Roentgen received the Nobel Prize for his discovery of X-rays, which had a significant impact on the field of physics and medicine\\\\Analysis:\\The facts presented in the statement you provided are largely accurate.\\\\Credibility Score: 10\\\#\#\#\#\#\#\#\#\#\#\\\\Passage:\\\{Passage\}
\end{tcolorbox}
\caption{Prompt for GPT to generate credibility scores.}
\label{fig:prompt_score}
\end{figure*}

\end{document}